# Context Matters: Recovering Human Semantic Structure from Machine Learning Analysis of Large-Scale Text Corpora

*Marius Cătălin Iordan\*, Tyler Giallanza, Cameron T. Ellis, Nicole M. Beckage,
Jonathan D. Cohen*

## Contact Information

Marius Cătălin Iordan, Ph.D.
*Princeton Neuroscience Institute & Psychology Department, Princeton University*
*Email: mci@princeton.edu*
*\*corresponding author*

Tyler Giallanza
Princeton Neuroscience Institute & Psychology Department, Princeton University
*Email: tylerg@princeton.edu*

Cameron T. Ellis
*Psychology Department, Yale University*
*Email: cameron.ellis@yale.edu*

Nicole M. Beckage, Ph.D.
*Intel Labs, Hillsboro, Oregon*
*Email: nicole.beckage@intel.com*

Jonathan D. Cohen, MD, Ph.D.
*Princeton Neuroscience Institute & Psychology Department, Princeton University*
*Email: jdc@princeton.edu*

## Manuscript Information

Number of Words
    Abstract      140
    Statement of Relevance      112
    Introduction & Discussion      1,979
    Methods & Results (incl. Figure Captions)      7,870
Number of References      68
Number of Display Items      5

**Abstract**

*Applying machine learning algorithms to large-scale, text-based corpora (embeddings) presents a unique opportunity to investigate at scale how human semantic knowledge is organized and how people use it to judge fundamental relationships, such as similarity between concepts. However, efforts to date have shown a substantial discrepancy between algorithm predictions and empirical judgments. Here, we introduce a novel approach of generating embeddings motivated by the psychological theory that semantic context plays a critical role in human judgments. Specifically, we train state-of-the-art machine learning algorithms using contextually-constrained text corpora and show that this greatly improves predictions of similarity judgments and feature ratings. By improving the correspondence between representations derived using embeddings generated by machine learning methods and empirical measurements of human judgments, the approach we describe helps advance the use of large-scale text corpora to understand the structure of human semantic representations.*

**Statement of Relevance**

*We describe advances in the use of machine learning methods for studying human semantic structure and judgments. Our method exceeds 90% of maximum achievable performance in automated out-of-sample prediction of empirical similarity judgments, as well as the best performance to date in predicting feature ratings (e.g., size) for concrete real-world objects (e.g., 'bear'). Our work also shows that it is possible to use large-scale text corpora to study the structure of semantic representations, complementing traditional laboratory-based approaches. It also provides a clear illustration that qualitative, psychologically relevant factors may be as important as the sheer quantity of data in constructing training sets for use with machine learning methods of investigating cognitive phenomena.*

**Introduction**

Understanding the underlying structure of human semantic representations is a fundamental and longstanding goal of cognitive science[1,2,3,4,5,6,7], with implications that range broadly from neuroscience[8,9] to computer science[10,11,12,13] and beyond[14]. Most theories of semantic knowledge propose that items in semantic memory are represented in a multidimensional feature space, and that key relationships among

items – such as similarity and category structure – are determined by distance among items in this space[1,2,4,15,16,17,18,19]. However, defining the space, establishing how distances are quantified, and using these distances to predict human judgments about semantic relationships such as similarity remains a challenge. Understanding similarity judgments is critical for providing insight into the structure and organization of human semantic knowledge, as these judgments provide one direct means of determining that structure. Furthermore, similarity judgments play a fundamental role in perception and reasoning, helping us to learn how new stimuli relate to previously learned categories, and to generalize this learning to novel situations. More specifically, similarity provides a key metric for a wide variety of cognitive processes such as categorization, identification, and prediction[1,4,15,16,18]. In the present work, we put forward a novel method for accurately estimating similarity relationships and item feature ratings from large-scale online word corpora as a crucial first step towards allowing complex cognitive processes to be studied and explained at scale.

The best efforts to define theoretical principles (e.g., formal metrics) that can predict semantic similarity judgments from empirical feature representations[3,16,20,21,22,23] capture less than half the variance observed in empirical studies of such judgments. At the same time, a comprehensive empirical determination of the structure of human semantic representation (e.g., by evaluating all possible similarity relationships) is impossible, given that human semantic experience encompasses billions of individual objects (e.g., millions of pencils, thousands of tables, all different from one another) and tens of thousands of categories[24] (e.g., 'pencil', 'table', etc.). Complementing research in cognitive psychology, natural language processing (NLP) has attempted to use large amounts of human generated text (billions of words[13,25,26,27]) to create a high dimensional representation that may provide insights into human semantic space. These approaches generate multidimensional vector spaces learned from the statistics of the input data, in which words that appear together across different sources of writing (e.g., articles, books) are evaluated as close to one another, and words that co-occur less are placed farther apart. A distance metric between a given pair of words can then be used as a measure of their similarity. This approach has met with some success in predicting categorical distinctions[28], predicting properties of objects[29,30,31], and even revealing cultural stereotypes and implicit associations hidden within the documents[14]. However, the spaces generated by such machine learning methods have remained limited in their ability to predict direct empirical measurements of human similarity judgments[10,29] and feature ratings[30].

Thus, neither the top-down theoretically-principled approaches, nor bottom-up data-driven approaches have provided an empirically-validated method of accurately and consistently predicting how humans judge similarity relationships or feature ratings between objects and concepts. In the present work, we put forward a novel method for accurately estimating these properties or concrete objects from large text corpora, as a crucial first step towards allowing such complex cognitive processes to be studied and explained at scale. Our approach is motivated by the hypothesis that the human semantic judgments are influenced not only by the local context of words in a text corpus, but also by the domain-level semantic context in which these judgments are made across many cognitive tasks[32,33,34,35,36,37], including when evaluating similarity relationships[38,39,40,41,42] (also see Supplementary Experiments 1–4 & Supplementary Fig. 1). This attentional influence can include task demands (e.g., instructions provided by experimenters), incidental factors related to the circumstances of the task, and/or features of the items to be judged. For example, when asked to judge the similarity between a bear and a bull among a number of other animals, attention may be directed to their physical characteristics as objects in a natural context (e.g., size), leading to the judgment that they are similar; however, in the context of financial markets, attention may be drawn to their economic value, leading to the judgment that they are very different. Current state-of-the-art NLP models show that taking local contextual influences into account (i.e., the 10-20 words that surround a concept) can improve performance on tasks such as question answering and ambiguous pronoun comprehension[43,44,45]. Here, we consider the question of how also taking into account of global, domain-level semantic context (the topic or domain being considered in the writings, e.g., National Geographic vs. Wall Street Journal) may impact the performance of machine learning models in predicting human similarity judgments and feature ratings.

To address this question, we modified the use of modern machine learning methods for generating data-driven high-dimensional semantic embedding spaces by introducing domain-level semantic contextual constraints in the construction of the text corpora from which the embedding spaces are learned. These constraints are intended to parallel the contextual constraints that might impact human judgments by restricting the corpora to materials generated by individuals (e.g., authors/speakers) whose mental context aligns more closely with that of the participants whose judgments are measured in empirical studies. We tested this approach in three experiments.

The first two experiments demonstrate that embedding spaces learned from contextually-constrained text corpora substantially improve the ability to predict empirical measures of human semantic judgments (pairwise similarity judgments in Experiment 1 and item-specific feature ratings in Experiment 2), despite being trained using two orders of magnitude *less* data than state-of-the-art natural language processing models[13,25,26,27,44]. In the third experiment, we test the usefulness of contextually-relevant features for generating distance metrics in context-free embedding spaces that can better recover empirical similarity judgments from such unconstrained spaces, though they still cannot match the performance of contextually-constrained corpora. Finally, we show that applying a similar method to embeddings from contextually-constrained corpora provides the best prediction of human similarity judgments achieved to date, exceeding 90% of human inter-rater reliability in two specific semantic contexts.

**Results**

*Experiment 1a: Contextually-Constrained Semantic Embeddings Capture Human Similarity Judgments Better than Context-Free Embeddings*

     Word embedding spaces are generated by training machine learning models on large corpora of text, often using deep neural network algorithms. This approach is typically applied to the largest corpora available, on the assumption that larger datasets will provide more accurate estimation of the underlying semantic structure. However, aggregating across multiple domain-level semantic contexts (e.g., National Geographic and Wall Street Journal) may dilute the sensitivity of resulting embedding spaces to contextually-constrained human semantic judgments. To test whether contextually-constraining the corpora used to train machine learning algorithms to particular domains improves their ability to predict empirical similarity judgments, we collected Wikipedia articles related to two distinct domain-level semantic contexts (Fig. 1a): 'nature' (~70 million words) and 'transportation' (~50 million words). For convenience, we henceforth refer to the models we built as 'contextually-constrained' since they take into account both local (word- and sentence-level) and global (domain- and discourse-level) context and we refer to existing state-of-the-art NLP models (trained by us or pre-trained) as 'context-free' to emphasize that they focus exclusively on local context during training. We generated contextually-constrained embedding spaces (CC nature and CC transportation) by training continuous skip-gram Word2Vec models with negative sampling[25,26] using the two collections of contextually-relevant articles (nature and transportation) as training sets. We chose Word2Vec to train our embedding spaces because this type of model has been shown to be superior to other embedding models in matching human semantic judgments (e.g., analogy) and requires relatively little data to train effectively[29].

    We compared the two contextually-constrained embedding spaces (CC nature and CC transportation) to a context-free Word2Vec embedding space trained on all English language Wikipedia articles (~2 billion words) and to an embedding space

trained on a random subset of this training corpus, size-matched to the contextually-constrained embedding spaces (~60 million words). We also compared the performance of our contextually-constrained embedding spaces to that of BERT[44] (trained on English language Wikipedia and English Books, ~3 billion words), a state-of-the-art transformer neural network pre-trained to take into account local context (i.e., the 10-20 words that surround a particular concept in a corpus of text). This network architecture has been shown to outperform Word2Vec[25] embeddings on multiple human-related cognitive tasks involving prediction of semantic knowledge, but has not been previously tested on its ability to match human similarity judgments. For consistency, we henceforth refer to this architecture as context-free, as well, as it does not take account of domain-level knowledge in how it is trained. Finally, we compared performance of the four Word2Vec embedding spaces to another commonly used context-free embedding space known as GloVe[27] for two main reasons: first, the GloVe embeddings are learned from the enormous Common Crawl corpus (~42 billion words) and thus provide insight into the role of corpus size on making predictions about human judgments; and, second, GloVe is a widely used independent algorithm for learning embeddings that does not rely on deep learning architectures, allowing us to evaluate the generality of our results beyond neural network embeddings.

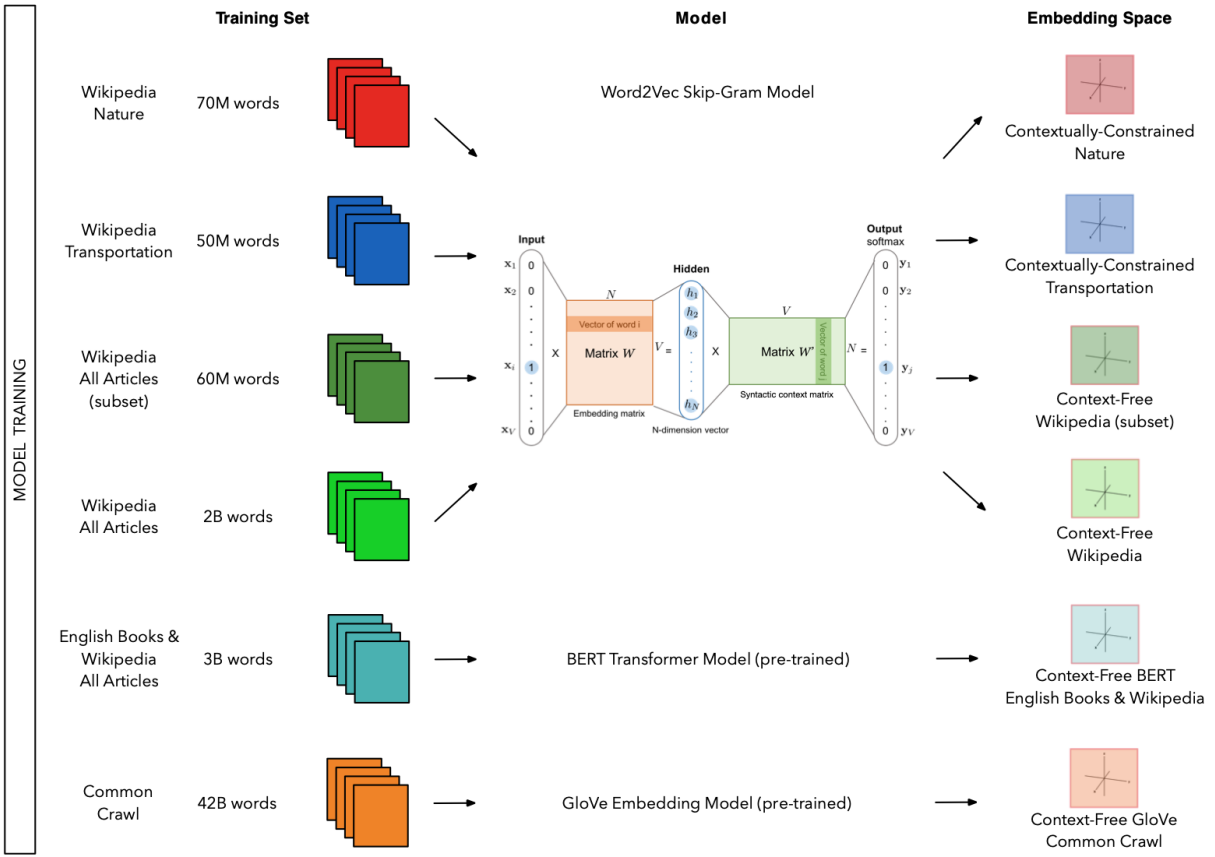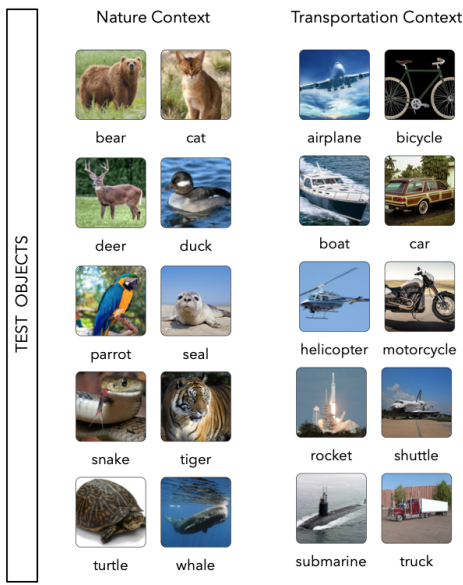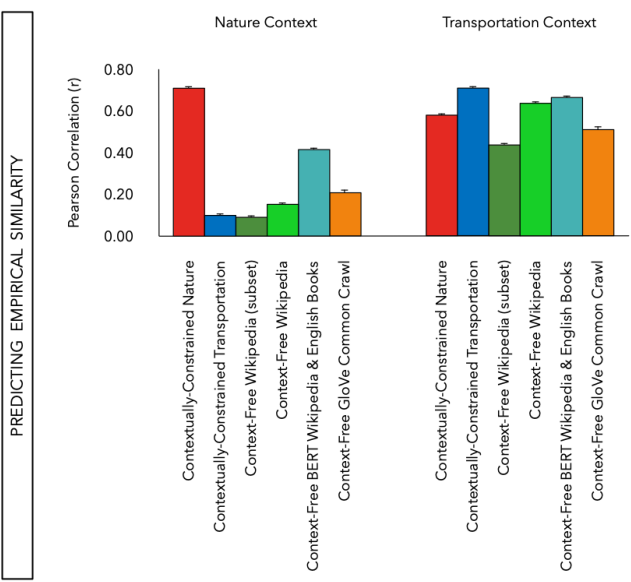

*Figure 1. Generating Contextually-Constrained Embedding Spaces and Testing Their Alignment with Human Similarity Judgments.* (a) Model Training. We generated contextually-constrained embedding spaces using training sets comprised of Wikipedia articles considered relevant to each semantic context ('nature' ~70M words, 'transportation' ~ 50M words). Similarly, we trained context-free models with the training set of all publicly available Wikipedia articles (~2B words), as well as a size-matched subset of this corpus (~60M words). We compared results using these models to a pre-trained BERT transformer network trained on Wikipedia articles and the English Books corpus (~3B words) and against GloVe, a pre-trained embedding space trained on the Common Crawl corpus (~42B words). (b) To quantify how well each embedding space aligned with human similarity judgments for the two semantic test contexts, we selected 10 representative basic-level objects for each test context (10 animals and 10 vehicles) and collected human-reported similarity judgments between all pairs of objects in each context (45 pairs per context). (c) We computed Pearson correlation between human empirical similarity judgments (all 45 pairwise comparisons within each semantic context, averaged across participants) and similarity predicted by each embedding model (cosine distance between embedding vectors corresponding to each object in each model). Error bars show 95% confidence intervals for 1,000 bootstrapped samples of the test-set items (see Methods for details). All differences between contextually-constrained models in their preferred context and other models are statistically significant, $p<=0.004$.

To test how well each embedding space could predict human similarity judgments, we selected two representative subsets of ten concrete basic-level objects[46] commonly associated with each of the two contexts (10 animals, e.g., 'bear'; 10 vehicles, e.g., 'car'; Fig. 1b). We chose common basic-level nouns used in prior work[3,23,39] because the age of acquisition, categorization, recognition speed, and multiple other cognitive and neural features have been shown to be fairly similar across items at this level of the semantic hierarchy[46,47,48,49]. We used the Amazon Mechanical Turk online platform to collect empirical similarity judgments on a Likert scale (1–5) for all pairs of objects within each context (e.g., by asking participants to judge "How similar are a bear and a cat?"; or "How similar are a bicycle and a car?"). Each participant made judgments in a single semantic context (i.e., all between-animal comparisons or all between-vehicle comparisons, but not both) and the order of comparisons in each context was counterbalanced across participants. After removing subject responses with low inter-rater reliability, we computed the empirical ground truth similarity for each pair of objects in each context as the average of these judgments across the remaining participants. We used cosine distance between embedding vectors corresponding to the 10 animals and 10 vehicles to generate similarity predictions for all pairwise combinations of objects in each of the embedding spaces. Finally, we calculated the Pearson correlation between the cosine distance measures and the empirical similarity judgments to assess how well each embedding space can account for human judgments of pairwise similarity.

For animals, estimates of similarity using the contextually-constrained nature embedding space were highly correlated with human judgments (CC nature r=0.711±0.004; Fig. 1c). By contrast, estimates from the contextually-constrained transportation embedding space and the context-free models could not recover the same pattern of human similarity judgments among animals (CC transportation r=0.100±0.003; Wikipedia subset r=0.090±0.006; Wikipedia r=0.152±0.008; Common Crawl r=0.207±0.009; BERT r=0.416±0.012; CC nature > CC transportation p<0.001; CC nature > Wikipedia subset p<0.001; CC nature > Wikipedia p<0.001; nature > Common Crawl p<0.001; CC nature > BERT p<0.001). Conversely, for vehicles, similarity estimates from its corresponding contextually-constrained transportation embedding space were the most highly correlated with human judgments (CC transportation r=0.710±0.009). While similarity estimates from the other embedding spaces were also highly correlated with empirical judgments of similarity among vehicles (CC nature r=0.580±0.008; Wikipedia subset r=0.437±0.005; Wikipedia r=0.637±0.005; Common Crawl r=0.510±0.005; BERT r=0.665±0.003), the ability to predict human judgments was significantly weaker than for the contextually-constrained transportation embedding space (CC transportation > nature p<0.001; CC transportation > Wikipedia subset p<0.001; CC transportation > Wikipedia p=0.004; CC transportation > Common Crawl p<0.001; CC transportation > BERT p=0.001). For both nature and transportation contexts, we observed that the state-of-the-art context-free BERT model performed approximately half-way between the context-free Wikipedia model and our embedding spaces that should be sensitive to the effects of both local and global context. The fact that our models consistently outperformed BERT in both semantic contexts suggests that taking account of domain-level semantic context in the construction of embedding spaces provides a more sensitive proxy for the presumed effects of semantic context on human similarity judgments than relying exclusively on local context (i.e., the surrounding words and/or sentences), as is the practice with existing NLP models.

This was further evidenced in two major ways: first, each contextually-constrained embedding space we built made more accurate predictions about judgments for its corresponding category of objects compared to the context-free models (e.g., using embeddings built from articles corresponding to the nature context to predict similarity judgments about animals); second, we observed a double dissociation between the performance of the contextually-constrained models according to context: predictions of similarity judgments were most substantially improved by using contextually-constrained corpora specifically when the contextual constraint aligned with the category of objects being judged, but these contextually-constrained representations did not generalize to other contexts. Furthermore, this double dissociation was robust across multiple hyper-parameter choices for the Word2Vec model, such as number of words surrounding a particular concept during training (i.e., window size), the dimensionality of the learned embedding spaces (Supplementary Figs. 2 & 3), and the number of independent initializations of the embedding models' training procedure (Supplementary Fig. 4). Moreover, all results we reported involved bootstrap sampling of the test set pairwise comparisons (see Methods for details), indicating that the difference in performance between models was reliable across item choices (i.e., particular animals or vehicles chosen for the test set). Finally, the results were robust to the choice of correlation metric used (Pearson vs. Spearman, Supplementary Fig. 5) and we did not observe any obvious trends in the errors made by networks and/or their agreement with human similarity judgments in the similarity matrices derived from empirical data or model predictions (Supplementary Fig. 6).

Together, these results strongly support the hypothesis that human similarity judgments can be better predicted by incorporating domain-level contextual constraints into the training procedure used to build word embedding spaces. Although the performance of the two contextually-constrained embedding models on their respective test sets was not equal, the difference cannot be explained by lexical features such as the number of possible meanings assigned to the test words (Oxford English Dictionary[50], WordNet[51]; Supplementary Table 1) or by the word frequency and absolute number of test words appearing in the training corpora (Supplementary Fig. 7 & Supplementary Table 2). However, a trend in WordNet meanings towards greater polysemy for animals versus vehicles may help partially explain why all models (contextually-constrained and context-free) were able to better predict human similarity judgments in the transportation context.

*Experiment 1b: Combining Semantic Contexts Degrades Contextually-Relevant Information*

We hypothesized that contextually-constrained embedding spaces would be better aligned with human judgments because they were built from corpora that reflect the influences of semantic contexts in the minds of the empirical study participants as they made similarity judgments. To further test this idea, we evaluated the extent to which cross-contextual contamination induces a misalignment between distances in embedding spaces and human similarity judgments. We generated new combined-context embedding spaces using different proportions of the training data from each of the two semantic contexts (nature and transportation; Fig. 2a), both matching for the size of the contextually-constrained models' training set (60M words; canonical combined-context model), as well as using all available training data from the two semantic contexts (120M words; full context-combined model).

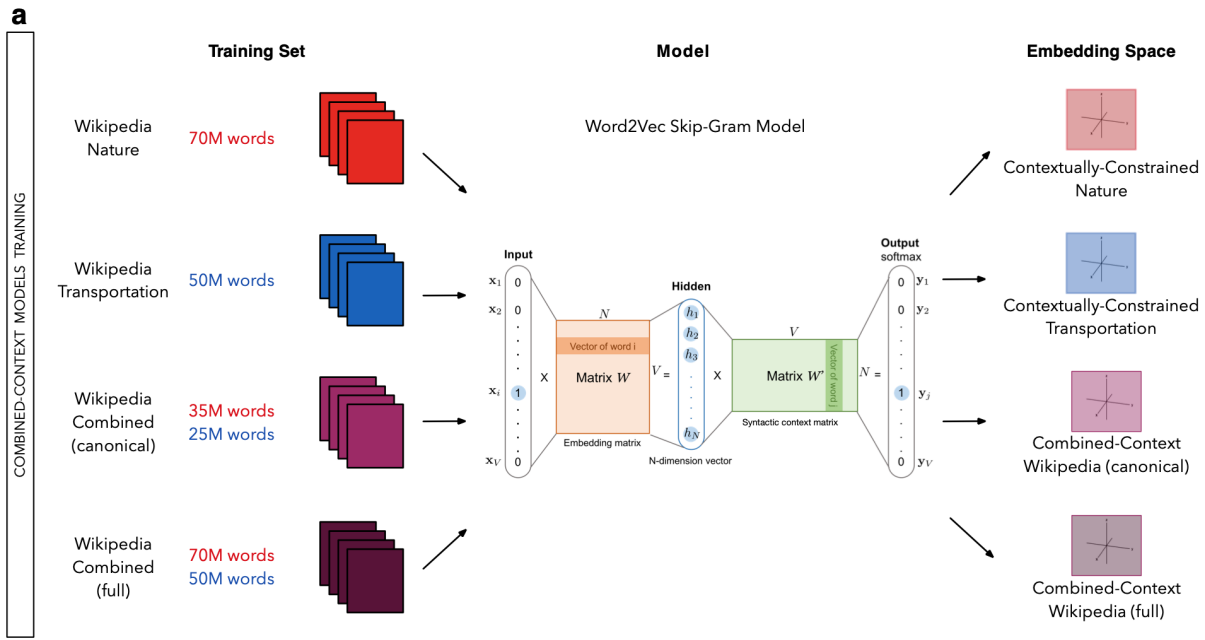

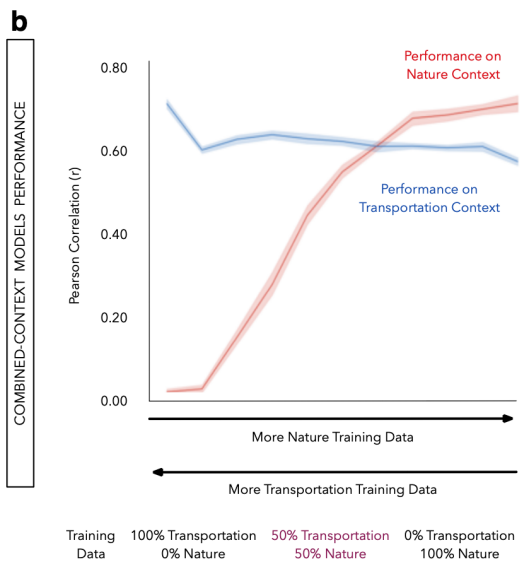

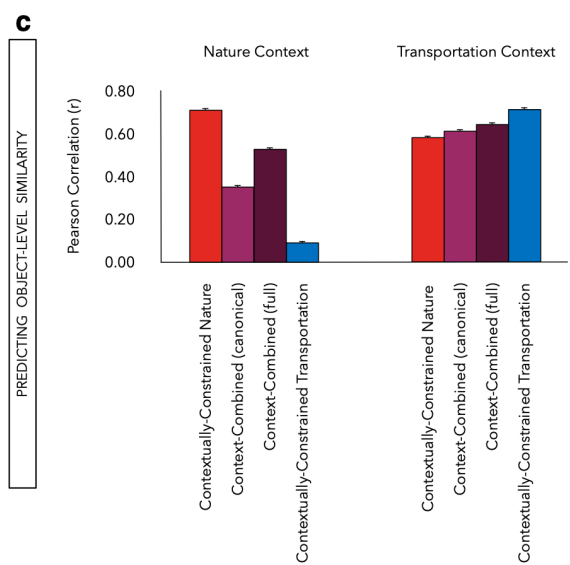

***Figure 2. Context-Combined Models More Poorly Predict Similarity Judgments.***
*(a) Combined-context embedding spaces were generated by using training data from the nature and transportation semantic contexts in different proportions (60M words, e.g., 10%-90%, 50%-50%, etc.). A full context-combined embedding space was also generated using all available training data from both semantic contexts (120M words). (b) When combining training data from two semantic contexts at different ratios (in increments of 10% training data for each context, e.g., 10% nature and 90 transportation, 20% nature and 80% transportation, etc.), the resulting combined-context embeddings recover a proportional amount of information from their preferred/non-preferred semantic contexts. (c) The canonical and full combined-context models produced distances between concepts that were less aligned with human judgments in both the nature and the transportation semantic contexts, respectively, compared to the corresponding contextually-constrained embedding spaces. Errors signify 95% confidence intervals 1,000 bootstrapped samples of the test-set items (see Methods for details).*

     As predicted, we found that performance of the combined-context embedding spaces was intermediate between the preferred and non-preferred contextually-constrained embedding spaces in predicting human similarity judgments: as more nature semantic context data was used to train the combined-context models, the alignment between embedding spaces and human judgments for the animal test set improved; and, conversely, more transportation semantic context data yielded better recovery of similarity relationships in the vehicle test set (Fig. 2b). We illustrated this performance difference using the 50% nature-50% transportation embedding spaces in Fig. 2c, but we observed the same general trend regardless of the ratios (nature context: combined canonical $r=0.354\pm0.004$; combined canonical < CC nature $p<0.001$; combined canonical > CC transportation $p<0.001$; combined full $r=0.527\pm0.007$; combined full < CC nature $p<0.001$; combined full > CC transportation $p<0.001$; transportation context: combined canonical $r=0.613\pm0.008$; combined canonical > CC nature $p=0.069$; combined canonical < CC transportation $p=0.008$; combined full $r=0.640\pm0.006$; combined full > CC nature $p=0.024$; combined full < CC transportation $p=0.001$).

     Crucially, we observed that when using all training examples from one semantic context (e.g., nature, 70M words) and adding new examples from a different context (e.g., transportation, 50M additional words), the resulting embedding space performed worse at predicting human similarity judgments than the contextually-constrained embedding space that used only half of the training data. This result strongly suggests that the contextual relevance of the training data used to generate embedding spaces can be more important than the amount of data itself. Contrary to

common practice, adding more training examples may, in fact, degrade performance if the extra training data are not contextually relevant to the relationships of interest (in this case, similarity judgments among items).

Together, these results suggest that combining training data from multiple semantic contexts when generating embedding spaces may be responsible in part for the misalignment between human semantic judgments and the relationships recovered by context-free embedding models (which are usually trained using data from many semantic contexts). This is consistent with an analogous trend observed when humans were asked to perform similarity judgments across multiple semantic contexts simultaneously (Supplementary Experiments 1–4 and Supplementary Fig. 1).

*Experiment 2: Contextually-Constrained Embeddings Capture More Information about Interpretable Object Features Compared to Context-Free Embeddings*

The findings above are consistent with the hypothesis that semantic judgments are subject to contextual (attentional) influences that constrain the scope of the knowledge representation (e.g., item features) used to make those judgments. However, the analyses presented so far provide little insight into the feature dimensions that define the underlying structure of semantic representations and on which similarity judgments are based. It is largely assumed that, for humans, feature dimensions are recognizable, simple ones, such as size, shape, location, function, etc.[16,23] However, it remains possible that the true underlying representation may instead be comprised of more abstract combinations of such simple features, and/or other perhaps uninterpretable features. From a machine learning perspective, embeddings based on large-scale, unconstrained corpora generally do not yield interpretable features, even when they generate results that capture some aspects of human performance[25,31]. One potential explanation for this may be that domain-level contextual constraints may play an important role in directing attention to particular features when these are being rated by humans, whereas this contextual influence is weakened when generating context-free embedding spaces (cf. Experiment 1b) and thus information along interpretable features may be less emphasized in such spaces, as well. To evaluate this possibility, we tested whether contextually-constrained embedding spaces would yield feature ratings for individual objects that are more closely aligned to humans on intuitively recognizable dimensions (e.g., size), as well as more relevant to predicting empirical similarity judgments.

To test how well embedding spaces could predict human feature ratings, we identified 12 context-relevant features for each of the two semantic contexts used in Experiment 1. For each semantic context, we collected six concrete features (nature: size, domesticity, predacity, speed, furriness, aquatic-ness; transportation: elevation, openness, size, speed, wheeledness, cost) and six subjective features (nature: dangerousness, edibility, intelligence, humanness, cuteness, interestingness; transportation: comfort, dangerousness, interest, personalness, usefulness, skill). The concrete features comprised a reasonable subset of features used throughout prior work on explaining similarity judgments, which are commonly listed by human participants when asked to describe concrete objects[3,46]. However, little data has been collected how well subjective (and potentially more abstract or relational[39,52]) features can predict similarity judgments between pairs of real-world objects. Our prior work has shown that such subjective features for the nature domain can capture more variance in human judgments, compared to concrete features[23]. Here, we extended this approach to identifying six subjective features for the transportation domain. We then used the Amazon Mechanical Turk platform to collect ratings of those features for the 10 test objects in each associated context; that is, the 10 animals were rated on the 12 nature features and the 10 vehicles were rated on the 12 transportation features (Likert scales 1–5 were used for all features and objects). A full list of features for each semantic context is given in Supplementary Tables 3 & 4.

To generate feature ratings from embedding spaces, we used a "contextual semantic projection" approach. In this method, for a given feature (e.g., size), a set of three "anchor" objects was chosen that corresponded to the low end of the feature range (e.g., 'bird, 'rabbit, 'rat') and a second set of three anchor objects was chosen that corresponded to the high end of the feature range (e.g., 'lion, 'giraffe, 'elephant'), none of which were in the test set. The embedding vector for each anchor object at one end was then subtracted from the embedding vector of each anchor object at the other end (9 differences total). These vector differences were averaged to create a one-dimensional subspace (e.g., "size" line) within the original word embedding space. Test words/concepts (e.g., 'bear') were projected onto that line and the relative distance between each word and the low-/high-end object represented a feature rating prediction for that word/concept. As noted, to ensure generality and avoid overfitting, the anchor objects were out-of-sample (i.e., distinct from the 10 test objects used for each semantic context), and were chosen as reasonable representatives of the low/high value on their corresponding feature.

Crucially, by selecting different endpoints for features common across the two semantic contexts (e.g., 'size'), this method allowed us to make feature ratings predictions in a manner specific to a particular semantic context (nature / transportation). For example, in the nature context, 'size' was measured as the vector from 'rat', 'rabbit', etc. to 'elephant', 'giraffe', etc. (*animals* in the training, but not in the testing set) and in the transportation context as the vector from 'skateboard', 'scooter', etc. to 'spaceship', 'carrier', etc. (*vehicles* not in the testing set). By contrast, prior work using projection techniques to extract feature ratings from embedding spaces[30] have used adjectives as endpoints, ignoring the potential influence of domain-level semantic context on similarity judgments (e.g., 'size' was defined as a vector from 'small', 'tiny', 'minuscule' to 'large', 'huge', 'giant', regardless of semantic context). However, as we argued above, feature ratings may be impacted by semantic context much as – and perhaps for the same reasons as – similarity judgments (e.g., the size of an ant may be judged to be small among animals, but to be enormous compared to an atom). To additionally test this hypothesis, we compared our contextual projection technique to the adjective projection technique with regard to its ability to consistently predict empirical feature ratings. A complete list of the contextual and adjective projection endpoints used for each semantic context and each feature are listed in Supplementary Tables 5 & 6.

We found that both projection techniques were able to predict human feature ratings with positive correlation values, suggesting that feature information can be recovered from embedding spaces via projection (Fig. 3 & Supplementary Fig. 8). Overall, contextual projection predicted human feature ratings much more reliably than adjective projection on 18 out of 24 features and was tied for best performance for an additional 5 out of 24 features. Adjective projection performed best on a single nature feature (dangerousness in the nature context). Furthermore, across both semantic contexts, using contextually-constrained embedding spaces (with either projection method), we were able to predict human feature ratings better than using context-free embedding spaces for 13 out of 24 features and were tied for best performance for an additional 9 out of 24 features. Context-free embeddings performed best on only 2 nature context features (cuteness and dangerousness). Finally, we observed that all models were able to predict empirical ratings somewhat better on concrete features (average r=0.570) compared to subjective features (average r=0.517). This trend was somewhat enhanced for contextually-constrained embedding spaces (concrete feature average r=0.663, subjective feature average r=0.530). This suggests that concrete features may be more easily captured and encoded by automated methods (e.g., embedding spaces), compared to subjective features, despite the latter likely playing a significant role in how humans evaluate similarity judgments[23]. Finally, our results were not sensitive to the initialization conditions of the embedding models used for predicting feature ratings or item-level effects (Supplementary Fig. 8 includes 95% confidence intervals for 10 independent initializations of each model and 1000 bootstrapped samples of the test-set items per model). Together, our results suggest that contextually-constrained embedding spaces, when used in conjunction with contextual projection, were the most consistent and accurate in their ability to predict human feature ratings compared to using context-free embedding spaces and/or adjective projection.

## Nature Context

| Dimension / Model | Aquatic | Cute | Dangerous | Domestic | Edible | Furry | Human | Intelligent | Interesting | Predatory | Size | Speed |
|---|---|---|---|---|---|---|---|---|---|---|---|---|
| Contextual Projection | | | | | | | | | | | | |
| Nature | **.90** | .55 | .58 | **.66** | **.75** | **.69** | **.72** | **.62** | .48 | **.69** | **.93** | .60 |
| Transportation | .77 | .46 | .53 | .31 | .45 | .21 | .30 | .34 | **.50** | .49 | .46 | **.61** |
| Wikipedia (full) | .74 | **.77** | .52 | .40 | .72 | .63 | .49 | .29 | .32 | .60 | .60 | **.63** |
| Adjective Projection | | | | | | | | | | | | |
| Nature | .81 | .51 | .37 | **.66** | .56 | .61 | .48 | .36 | .41 | .56 | .56 | .51 |
| Transportation | .66 | .26 | .36 | .50 | .48 | .29 | .28 | **.60** | **.49** | .41 | .41 | .29 |
| Wikipedia (full) | .46 | .48 | **.68** | .33 | .22 | .26 | .25 | .37 | .46 | .46 | .47 | .29 |

## Transportation Context

| Dimension / Model | Comfort | Cost | Dangerous | Elevation | Interest | Open | Personal | Size | Skill | Speed | Useful | Wheeled |
|---|---|---|---|---|---|---|---|---|---|---|---|---|
| Contextual Projection | | | | | | | | | | | | |
| Nature | .37 | .89 | .80 | .61 | **.81** | **.78** | **.88** | **.82** | .85 | .59 | .36 | .50 |
| Transportation | **.48** | **.94** | **.84** | **.75** | **.80** | **.81** | .71 | **.85** | **.90** | **.71** | **.39** | **.77** |
| Wikipedia (full) | .41 | .92 | .72 | .59 | .77 | .70 | **.89** | .81 | **.88** | .54 | .36 | .73 |
| Adjective Projection | | | | | | | | | | | | |
| Nature | **.48** | .45 | .34 | .40 | .51 | .28 | .51 | .49 | .64 | .31 | .28 | .73 |
| Transportation | .35 | .52 | .35 | .32 | .27 | .31 | .48 | .36 | .45 | .65 | .33 | **.75** |
| Wikipedia (full) | .27 | .48 | .61 | .35 | .41 | .35 | .57 | .33 | .80 | .55 | .32 | .71 |

*Fig. 3. Contextual Projection Recovers Human Feature Ratings. Pearson correlations between predicted feature ratings using the contextual and adjective projection methods for items in the nature context (animals) and items in the transportation context (vehicles) with empirically obtained human feature ratings for corresponding semantic contexts. Across both nature and transportation semantic contexts, using contextual projection generated ratings that were better aligned with human judgments compared to other models and projection methods for 18 out of the 24 features considered and tied for best for an additional 5 out of 24 features. Furthermore, contextually-constrained embeddings (using either projection method) predicted feature ratings best on 13 out of 24 features and were tied for best performance for an additional 9 out of 24 features. Significance testing was done using 10 independent initializations of the model training procedure and 1000 bootstrapped samples of the test-set items each (Supplementary Fig. 8). Bolding and highlights indicate best (or tied for best) performing model in each column (red – contextually-constrained nature; blue – contextually-constrained transportation; green – context-free).*

Similar projection procedures applied to the canonical combined-context embedding space generated in Experiment 1b (50% nature–50% transportation, 60M words) yielded feature information that was less well aligned with human feature ratings, compared to the original contextually-constrained models, but better aligned than the contextually-constrained models for the other context, or the context-free models (Fig. 4 & Supplementary Fig. 9).

### Nature Context

| Dimension / Model | Aquatic | Cute | Dangerous | Domestic | Edible | Furry | Human | Intelligent | Interesting | Predatory | Size | Speed |
|---|---|---|---|---|---|---|---|---|---|---|---|---|
| Contextual Projection | | | | | | | | | | | | |
| Nature | **.90** | .55 | **.58** | **.66** | **.75** | **.69** | .72 | **.62** | .48 | **.69** | **.93** | .60 |
| Combined | .80 | .52 | .52 | .29 | .70 | .62 | **.77** | **.60** | .45 | .58 | .65 | **.75** |
| Transportation | .77 | .46 | .53 | .31 | .45 | .21 | .30 | .34 | **.50** | .49 | .46 | .61 |
| Wikipedia (full) | .74 | **.77** | .52 | .40 | .72 | .63 | .49 | .29 | .32 | .60 | .60 | .63 |

### Transportation Context

| Dimension / Model | Comfort | Cost | Dangerous | Elevation | Interest | Open | Personal | Size | Skill | Speed | Useful | Wheeled |
|---|---|---|---|---|---|---|---|---|---|---|---|---|
| Contextual Projection | | | | | | | | | | | | |
| Nature | .37 | .89 | .80 | .61 | **.81** | **.78** | **.88** | **.82** | .85 | .59 | .36 | .50 |
| Combined | **.59** | .91 | .81 | .61 | **.81** | .70 | .86 | .81 | **.88** | **.70** | .35 | .74 |
| Transportation | .48 | **.94** | **.84** | **.75** | **.80** | **.81** | .71 | **.85** | **.90** | **.71** | **.39** | **.77** |
| Wikipedia (full) | .41 | .92 | .72 | .59 | .77 | .70 | **.89** | .81 | **.88** | .54 | .36 | .73 |

***Fig. 4. Combined-Context Embedding Spaces Recover Feature Ratings Less Well Than Contextually-Constrained Embedding Spaces.*** *Pearson correlation between predicted feature ratings using contextual projection applied to the canonical combined-context embedding space (50% nature – 50% transportation, 60M words) and empirical human feature ratings. Across both contexts, the contextually-constrained embeddings were best aligned with human judgments on 15 out of the 24 features considered, while the combined-context embeddings were best or tied for best for only 7 out of 24. Significance testing was done using 10 independent initializations of the model training procedure and 1000 bootstrapped samples of the test-set items each (Supplementary Fig. 9). Bolding and highlights indicate best (or tied for best) performing model in each column (red – contextually-constrained nature; purple – context-combined; blue – contextually-constrained transportation; green – context-free).*

Together, the findings of Experiments 1 and 2 support the hypothesis that contextually-constrained embedding spaces can predict human similarity judgments and features ratings more consistently than context-free embedding spaces in their respective semantic contexts. We also show that training embedding spaces on corpora that include multiple domain-level semantic contexts substantially degrades both the ability to predict object-level similarity judgments (Experiment 1), as well as object feature values (Experiment 2), even though these types of judgments are easy for humans to make and reliable across individuals.

*Experiment 3: Using Contextually-Relevant Features to Improve Prediction of Human Similarity Judgments from Context-Free Embeddings*

Context-free embeddings are built from large-scale corpora comprising billions of words that likely span hundreds of semantic contexts. Currently, such embedding spaces are a key component of many application domains, ranging from neuroscience[8,9] to computer science[10,11,12,13] and beyond[14]. Our work suggests that if the goal of these applications is to solve human-relevant problems, then at least some of these domains may benefit from employing contextually-constrained embedding spaces instead, which would better predict human semantic structure. However, re-training embedding models using different text corpora and/or collecting such domain-level semantically-relevant corpora on a case-by-case basis may be expensive or difficult in practice. To help alleviate this problem, we propose an alternative approach that uses contextually-relevant features to predict human semantic information (e.g., similarity judgments) from *context-free* embedding spaces.

Previous work in cognitive science has attempted to predict similarity judgments from object feature values by collecting empirical ratings for objects along different features and computing the distance (using various metrics) between those feature vectors for pairs of objects. Such methods consistently explain about a third of the variance observed in human similarity judgments[3,4,16,21,53]. They can be further improved using regression methods to differentially weight the dimensions, but at best this can only explain about half the variance in human similarity judgments (e.g., r=0.65[23]).

Here, we test the hypothesis that human similarity judgments can be better predicted from context-free embedding spaces by using contextually-relevant features (cf. Experiment 2) together with the regression methods employed in cognitive psychology experiments that attempt to predict similarity between objects based on such features. First, we used the contextual projection method described in Experiment 2 to construct 12-dimensional subspaces for each embedding space corresponding to the 12 features in each semantic context (see Supplementary Tables 3 & 4 for details on features and endpoints). Second, we used linear regression to learn optimal weights for each feature in the 12-dimensional subspace of each embedding space that together best predicted human similarity judgments. To evaluate this procedure, we performed cross-validated out of sample prediction by repeatedly selecting one of the ten test objects in each semantic context and learning regression weights (training) that best predict human similarity judgments using a given high-dimensional embedding (e.g., 12D, 100D) using only empirical trials that don't involve the left-out object (36 out of 45 trials per feature; 80% of the empirical data). Subsequently, we used the learned weights to make new predictions on the left-out 20% of the judgments (9 out of 45 trials per feature, each comparison between the left-out object and the other 9 test objects for that semantic context; Fig. 5).

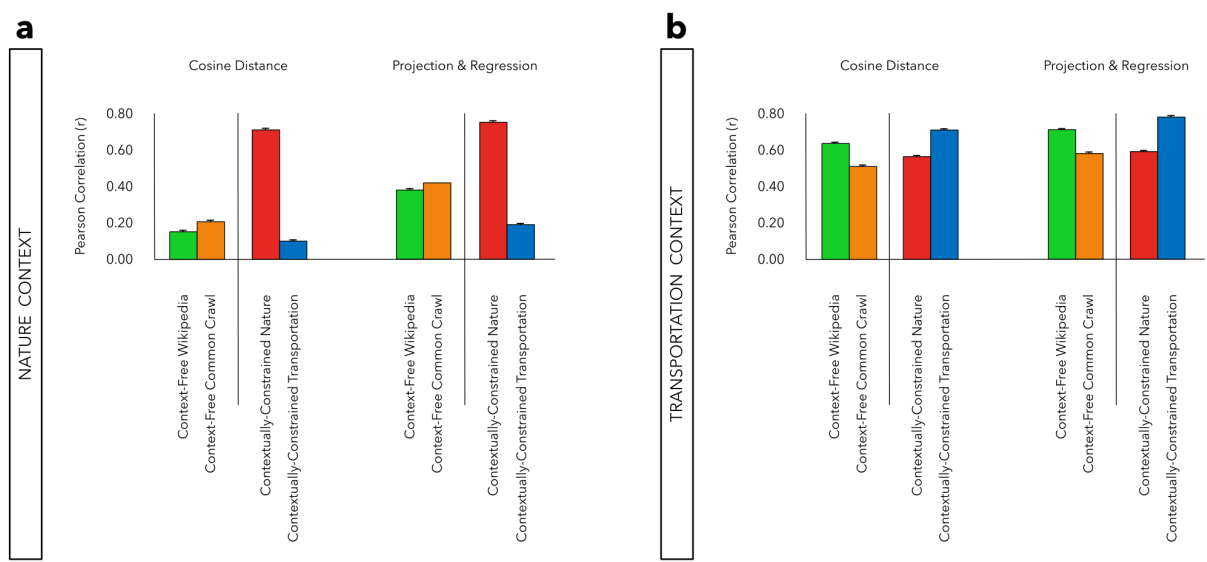

*Figure 5. Contextual Projection and Linear Regression Significantly Improve Recovery of Human Similarity Judgments from Embedding Spaces. Contextual projection was used to generate 12-dimensional subspaces for each embedding space corresponding to the 12 features for each semantic context. Linear regression was then used to learn optimal weights for each feature in each subspace that together best predicted human similarity judgments. Graphs show Pearson correlation between out-of-sample cross-validated predicted similarity values using the projection and regression procedure and human similarity judgments. (a) Nature semantic context. (b) Transportation semantic context. Error bars show 95% confidence intervals for 1,000 total bootstrapped samples of the test-set items (see Methods for details). All differences between 'projection and regression' bars and corresponding 'cosine' bars are statistically significant, $p<=0.020$.*

The contextual projection and regression procedure significantly improved predictions of human similarity judgments for all context-free embedding spaces (Fig. 5; nature context, projection & regression > cosine: Wikipedia $p<0.001$; Common

Crawl p<0.001; transportation context, projection & regression > cosine: Wikipedia p<0.001; Common Crawl p=0.008). By comparison, neither learning weights on the original set of 100 dimensions in each embedding space via regression (Supplementary Fig. 10), nor using cosine distance in the 12-dimensional contextual projection space, which is equivalent to assigning the same weight to each feature (Supplementary Fig. 11), could predict human similarity judgments as well as using both contextual projection and regression together. These results suggest that the improved accuracy of combined contextual projection and regression provide a novel and more accurate approach for recovering human-aligned semantic relationships that appear to be present, but previously inaccessible, within context-free embedding spaces.

Finally, if people differentially weight different dimensions when making similarity judgments, then the contextual projection and regression procedure should also improve predictions of human similarity judgments from contextually constrained embeddings. Our findings not only confirm this prediction (Fig. 5; nature context, projection & regression > cosine: CC nature p=0.030, CC transportation p<0.001; transportation context, projection & regression > cosine: CC nature p=0.009, CC transportation p=0.020), but also provide the best prediction of human similarity judgments to date using either human feature ratings or text-based embedding spaces, with correlations of up to r=0.75 in the nature semantic context and up to r=0.78 in the transportation semantic context. This accounts for 92% and 90% of human inter-rater variability in human similarity judgments for these two contexts. Contextual projection with learned weights shows substantial improvement upon the best previous prediction of human similarity judgments using empirical human feature ratings (r=0.65[23]). Remarkably, the predictions are from features extracted from artificially-built word embedding spaces (not empirical human feature ratings), trained using two orders of magnitude less data that state-of-the-art NLP models (~50 million words vs. 2–42 billion words) and are evaluated using an out-of-sample prediction procedure. The ability to reach or exceed 90% of human inter-rater reliability in these specific semantic contexts suggests that this computational approach provides an accurate and robust representation of the structure of human semantic knowledge.

**Discussion**

Our results support the hypothesis that efforts to understand and automatically infer how semantic knowledge is organized can benefit by taking both domain-level semantic context and local context into account, whether in empirical studies or when using machine learning methods applied to large-scale text corpora. Moreover, we showed that such machine learning approaches can be used to reliably estimate human semantic similarity judgments and object feature ratings, provided that information relevant to the domain-level contextual effects of human attention is used, either in the construction of the training corpora (Experiments 1–2) or in the method used to make predictions about human judgments from embedding spaces (Experiment 3).

From a psychological and cognitive science perspective, discovering reliable mappings between data-driven approaches and human judgments may help improve long-standing models of human behavior for tasks such as categorization, learning, and prediction. Understanding how people carry out such tasks requires the ability to reliably estimate similarity between concepts, identify features that describe them, and characterize how attention impacts these measurements – efforts that, for practical reasons, have so far focused on either artificially-built examples (e.g., sets of abstract shapes) or small-scale subsets of cognitive space[2,3,21,23,34,54]. Our work strongly supports the claim that psychological domain-level semantic context is important for human judgments and that current state-of-the-art machine learning models trained on billions of words may benefit from taking domain-level information into account. Our approach will potentially allow us to study tasks that rely on similarity judgments (e.g., categorization) and the influence of attention on these judgments much more efficiently and at a much broader scale than currently possible in laboratory studies.

From a human neuroscience perspective, it is unlikely that humans retrain their long-standing semantic representations every time a new task demands it; instead, attention is thought to alter the context in which learned semantic structure is processed, both in behavior and in the brain[55,56,57,58]. Recent advances in neuroimaging have allowed embedding-based neural models of semantics to probe how concepts are processed across the human brain[8,59] and to generate decoders of mental representations that can predict human behavior from neural responses[9]. Increasing alignment between embedding spaces and human semantic structure helps further our understanding of the structural underpinnings of semantic knowledge[18,60]. As such, our results suggest a novel avenue for investigating the mechanisms of how context dynamically shifts human behavior and neural responses across large-scale semantic structure.

From a natural language processing (NLP) perspective, embedding spaces have been used extensively as a primary building block, under the assumption that these spaces represent adequate proxies for human syntactic and semantic structure. By substantially improving alignment of embeddings with empirical object feature ratings and similarity judgments, we introduce new tools for exploration of cognitive phenomena with NLP. Both human-aligned embedding spaces resulting from contextually-constrained training, and projections that are motivated and validated on empirical data, may lead to improvements in the performance of NLP models that rely on embedding spaces to make inferences about human decision making and task performance. Example applications include machine translation[10], automatic extension of knowledge bases[11], text summarization[12], and image and video captioning[61,62,63,64].

When using NLP (and, more broadly, machine learning) to investigate human semantic structure, it is generally assumed that increasing the size of the training corpus should increase performance[25,29]. However, our results suggest an important countervailing factor: the extent to which the training corpus reflects the influence of the same attentional factors (domain-level semantic context) as the subsequent testing regime. In our experiments, contextually-constrained models trained on corpora comprising 50-70 million words outperformed state-of-the-art context-free models trained on billions or tens of billions of words. This demonstrates that data quality (as measured by contextual relevance) may be just as important as data quantity in building embedding spaces that would capture relationships salient to the specific task in which such spaces are employed. This key finding may provide further avenues of exploration for researchers building data-driven artificial language models that aim to emulate human performance on a plethora of tasks.

Recently, new approaches have been proposed that aim to incorporate contextual influences into artificial language models, such as BERT[44], ELMO[43], and multi-sense embeddings[45]. Although such models can perform well on language tasks such as question answering, next sentence prediction, and ambiguous pronoun comprehension, the models only address the effects of local context (i.e., the 10-20 words that surround a particular concept). By comparison, our approach also encompasses the effects of global, discourse-level semantic effects (e.g., the topic or domain being considered in the writings), in addition to local contextual effects. For example, BERT[44] is considered state-of-the-art for the tasks listed above and outperformed context-free Word2Vec and GloVe embedding models at predicting empirical similarity judgments. However, it could not match the performance of our contextually-constrained models, despite using significantly more training data than our Word2Vec embedding models (3 billion vs. 50-70 million words, Fig. 1). This provides strong evidence that predictions of NLP models can be further improved by additionally taking account of global, discourse-level context, which is consistent with observations from studies in cognitive science over the past forty years[32,33,34,35,36,37,38,39,40,41,42]. This suggests that our approach complements that of state-of-the-art artificial language models and that it may be possible to combine the two approaches to improve the ability to predict human similarity judgments even further, a topic we identify as valuable to pursue in future work.

The method and observations we report strengthen the existing link between human semantic space (how we organize knowledge and use it to interact with the world) and machine learning methods meant to automate tasks useful and directly relevant to humans (e.g., NLP). The ability to reach or exceed 90% of maximum achievable performance in predicting similarity between concepts in these specific semantic contexts validates a set of computational tools that allow for more accurate and robust representations of human semantic knowledge and are likely to be helpful in understanding the underlying structure of human semantic representations and in efforts to build artificial systems that can emulate and/or better interact with semantic representations.

**Methods**

*Generating Word Embedding Spaces*

We generated all semantic embedding spaces using the continuous skip-gram Word2Vec model with negative sampling as proposed by Mikolov et al.[25,26], henceforth referred to as 'Word2Vec'. Word2Vec has been shown to be superior to, or on par with, other embedding models at matching human similarity judgments[29]. The assumption behind Word2Vec is that words that appear in similar syntactic contexts (i.e., surrounded by, co-occurring with, or in a window size of a similar set of 8–12 words) tend to have similar meanings. To encode this relationship between words, Word2Vec represents each word as a multidimensional vector. The algorithm preferentially learns word vectors of a specific dimension such that the ability of a vector representation of a word to predict other words within a given syntactic window is maximized (i.e., words from the same syntactic window are placed close to each other in this multidimensional space, as are words whose syntactic windows are similar to one another).

We trained four primary types of embedding spaces: (1) contextually-constrained models (CC 'nature' and CC 'transportation'); (2) context-combined models; and (3) context-free models. Contextually-constrained models (1) were trained on a subset of English language Wikipedia determined by human-curated category labels (meta-information available directly from Wikipedia) associated with each Wikipedia article. Each category contains multiple articles and multiple sub-categories; the categories of Wikipedia thus form a tree where articles themselves are the leaves. We constructed the 'nature' semantic context training corpus by traversing the tree rooted at the 'animal' category, and we constructed the 'transportation' semantic context training corpus by combining the trees rooted at the 'transport' and 'travel' categories. To avoid topics unrelated to natural semantic contexts, we removed the sub-tree 'humans' from the 'nature' training corpus. Furthermore, to ensure that the animal and vehicle contexts were non-overlapping, we removed training articles that were labeled as belonging to both the 'nature' and 'transportation' training corpora. This yielded final training corpora of approximately 70 million words for the 'nature' semantic context and 50 million words for the 'transportation' semantic context. The combined-context models (2) were trained by combining data from each of the two contextually-constrained training corpora in varying amounts. For the models that matched training corpora size with the contextually-constrained models, we selected proportions of the two corpora that added up to approximately 60 million words (e.g., 10% 'transportation' corpus + 90% 'nature' corpus). The canonical size-matched combined-context model was obtained using a 50%-50% split (i.e., approximately 35 million words from the 'nature' semantic context and 25 million words from the 'transportation' semantic context). We also trained a combined-context model that included all training data used to generate both the 'nature' and the 'transportation' contextually-constrained models (full combined-context model, approximately 120 million words). Finally, the context-free models (3) were trained using English language Wikipedia articles unrestricted to a particular category (or semantic context). The full context-free Wikipedia model was trained using the full corpus of text corresponding to all Wikipedia articles (approximately 2 billion words) and the size-matched context-free model was trained by sampling 60 million words from this full corpus.

The primary factors controlling the Word2Vec model are the size of the syntactic context window ('window size') and the dimensionality of the resulting embedding space. Larger window sized result in embedding spaces that capture relationships between words that are farther apart in a document, and larger dimensionality has the potential to represent more of these relationships between words in a vocabulary. In practice, as window size or vector length increase, larger amounts of training data are required. To build our embedding spaces, we first conducted a grid search of all window sizes in the set (8, 9, 10, 11, 12) and all dimensionalities in the set (100, 150, 200) and selected the combination of parameters which yielded the highest agreement between similarity predicted by the Wikipedia context-free model (4) (2 billion words) and empirically-collected human similarity judgments (see *Human Behavioral Experiments* section). We reasoned that this would provide the most stringent possible benchmark of the context-free embedding spaces for evaluating our contextually-constrained embedding spaces against. Accordingly, all results and figures in the manuscript are obtained using models with a window size of 9 words and a dimensionality of 100 (Supplementary Figs. 2 & 3).

All models were trained using the 'gensim' Python library's implementation of the Word2Vec model[65]. Aside from window size and dimensionality, all other parameters were kept as the default values from the original Word2Vec publications[25,26]: an initial learning rate of 0.025, elimination of words that appear fewer than 5 times in the training corpus, a 0.001 threshold for downsampling frequently occurring words, an exponent of 0.75 for shaping the negative sampling distribution, 5 negative samples per positive sample, and the skip-gram training algorithm. Given that the final value of the loss function optimized during training is not comparable across networks and/or across datasets/training corpora[25,26], we trained every network for a fixed number of iterations. The resulting vocabulary sizes for each embedding space we constructed were: 148K vectors for the contextually-constrained 'nature' model, 110K words for the contextually-constrained 'transportation' model, 204K words for the combined-context models (canonical & full), 342K words for the context-free Wikipedia full model, and 125K words for the context-free Wikipedia subset model.

For each type of model (contextually-constrained, combined-context, context-free), we trained ten separate models with different initializations (but identical hyper-parameters) to control for the influence that random initialization of the weights has on model performance. Cosine similarity was used as a distance metric between two learned word vectors. Subsequently, we averaged the similarity values obtained for the ten models into one aggregate mean value. For this mean similarity, we performed bootstrapped sampling[66] of all the object pairs with replacement to evaluate how stable the similarity values are given the choice of test objects (1000 total samples). We reported the mean and 95% confidence intervals of the full 1000 samples for each model evaluation[66].

We also compared against two pre-trained models: (a) the BERT transformer network[44] generated using a corpus of 3 billion words (English language Wikipedia and English Books corpus); and (b) the GloVe embedding space[27] generated using a corpus of 42 billion words (freely available online: https://nlp.stanford.edu/projects/glove/). The pre-trained GloVe model had a dimensionality of 300 and a vocabulary size of 400K words. For this model, we performed the sampling procedure detailed above 1000 times and reported the mean and 95% confidence intervals of the full 1000 samples for each model evaluation. The BERT model was pre-trained on a corpus of 3 billion words comprising all English language Wikipedia and the English books corpus. The BERT model had a dimensionality of 768 and a vocabulary size of 300K tokens (word-equivalents). For the BERT model, we generated similarity predictions for a pair of text objects (e.g., bear and cat) by selecting 100 pairs of random sentences from the training set, each containing one of the two test objects, and comparing the cosine distance between the resulting embeddings for the two words in the highest (last) layer of the transformer network (768 nodes). The average similarity across the 100 pairs represented one BERT 'model' (we did not retrain BERT). The procedure was then repeated 10 times, analogously to the 10 separate initializations for each of the Word2Vec models we built. Finally, similar to the contextually-constrained Word2Vec models, we averaged the similarity values obtained for the ten BERT 'models' and performed the bootstrapping procedure 1000 times and reported the mean and 95% confidence interval of the resulting similarity prediction for the 1000 total samples.

*Object and Feature Testing Sets*

To test how well the trained embedding spaces aligned with human cognitive judgments, we constructed a stimulus set comprising ten representative basic-level animals (bear, cat, deer, duck, parrot, seal, snake, tiger, turtle, and whale) for the nature semantic context and ten representative basic-level vehicles (airplane, bicycle, boat, car, helicopter, motorcycle, rocket, shuttle, submarine, truck) for the transportation semantic context (Fig. 1b). We also selected twelve human-relevant features independently for each semantic context which have been shown to explain object-level similarity judgments in empirical settings[3,23,67]. For each semantic context, we collected six concrete features (nature: size, domesticity, predacity, speed, furriness, aquatic-ness; transportation: elevation, openness, size, speed, wheeledness, cost) and six subjective features (nature: dangerousness, edibility, intelligence, humanness, cuteness, interestingness; transportation: comfort, dangerousness, interest, personalness, usefulness, skill). The concrete features comprised a reasonable subset of features used throughout prior work on explaining similarity judgments, which are commonly listed by human participants when asked to describe concrete objects[3,46]. However, little data has been collected how well subjective (and potentially more abstract or relational[39,52]) features can predict similarity judgments between pairs of real-world objects. Our prior work has shown that such subjective features for the nature domain can capture more variance in human judgments, compared to concrete features[23]. Here, we extended this approach to identifying six subjective features for the transportation domain (Supplementary Table 4).

For each of the twenty total object categories (e.g., bear (animal), airplane (vehicle)), we collected nine images showcasing the animal in its natural habitat or the vehicle in its normal domain of operation. All images were in color, featured the target object as the largest and most prominent object on the screen, and were cropped to a size of 500x500 pixels each (one representative image from each category shown in Fig. 1b).

*Human Behavioral Experiments*

To collect empirical similarity judgments, we recruited 139 participants (45 female, 108 right-handed, mean age 31.5 years) through the Amazon Mechanical Turk online platform in exchange for $1.50 payment (expected rate $7.50/hour). Prior work has shown that for this type of task inter-participant reliability should be high for a cohort of at least 20 participants[23]. Participants were asked to report the similarity between every pair of objects from a single semantic context (e.g., all pairwise combinations of ten vehicles or all pairwise combinations of ten animals) on a discrete scale of 1 to 5 (1 = not similar; 5 = very similar). In each trial, the participant was shown two randomly selected images from each category side-by-side and was given unlimited time to report a similarity judgment. Each participant made 45 comparisons (all pairwise combinations of 10 categories from a single randomly chosen semantic context) presented in a random order.

To ensure high-quality judgments, we limited participation only to Mechanical Turk workers who had previously completed at least 1000 HITs with an acceptance rate of 95% or above. We excluded 34 participants who had no variance across answers (e.g. choosing a similarity value of 1 for every object pair). Prior work has shown that for this type of task inter-participant reliability should be high[23], therefore, to exclude participants whose response may have been random, we correlated the responses of each participant with the average of the responses for every other participant and calculated the Pearson correlation coefficient. We then iteratively removed the participant with the lowest Pearson coefficient, stopping this procedure when all remaining participants had a Pearson coefficient greater than or equal to 0.5 to the rest of the group. This excluded an additional 12 participants, leading to a final tally of n=44 participants for the nature semantic context and n=49 participants for the transportation semantic context.

To collect empirical feature ratings, we recruited 915 participants (392 female, 549 right-handed, mean age 33.4 years) through the Amazon Mechanical Turk online platform in exchange for $0.50 payment (expected rate $7.50/hour). Prior work has shown that for this type of task inter-participant reliability should be high for a cohort of at least 20 participants per feature[23]. Participants were asked to rank every object from a single semantic context (e.g., all ten vehicles or all ten animals) along a randomly chosen, context-specific dimension (e.g., "How fast/slow is this vehicle?") on a discrete scale of 1 to 5 (1 = low feature value, e.g. 'slow'; 5 = high feature value, e.g. 'fast'). In each trial, the participant was shown three randomly selected images from a total of nine possible images representing the object, as well as the name of the object (e.g., 'bear') and given unlimited time to report a feature rating. Each participant ranked all ten objects, presented in a random order, from a single randomly chosen context along a single randomly chosen dimension.

We used an analogous procedure as in collecting empirical similarity judgments to select high-quality responses (e.g., restricting the experiment to high performing workers and excluding 210 participants with low variance responses and 124 participants with answers that correlated poorly with the average response). This resulted in 18–33 total participants per feature (see Supplementary Tables 3 & 4 for details).

All participants had normal or corrected-to-normal visual acuity and provided informed consent to a protocol approved by the Princeton University Institutional Review Board.

*Predicting Similarity Judgments from Embedding Spaces*

To predict similarity between two objects in an embedding space, we computed the cosine distance between the vector representations of each object. We used cosine distance as a metric for two main reasons. First, cosine distance is a commonly reported metric used in the literature that allows for direct comparison to previous work[25,26,27,28,29]. Second, cosine distance disregards the length or magnitude of the two vectors being compared, taking into account only the angle between the vectors. Some studies[68] have demonstrated a relationship between the frequency with which a word appears in the training corpus and the length of the word vector. Because this frequency relationship should not have any bearing on the semantic similarity of the two words, using a distance metric such as cosine distance that ignores magnitude/length information is prudent.

Additionally, we developed a novel method for predicting semantic similarity judgments from word embedding spaces by first projecting embeddings onto human-relevant features (see section on *Feature Projections in Embedding Spaces* below) and then using these feature ratings to determine semantic similarity. To do so, we used either adjective or contextual projection to generate ratings for objects along the twelve human-relevant features we selected for each semantic context and subsequently used linear regression to estimate an optimal weighting for the features in each embedding space that would yield the best prediction of human similarity judgments. We used a leave-one-object-out cross-validation procedure where all pairwise comparisons between 9 of the 10 objects in each semantic context were used to estimate the feature weights and then the weights were used to predict similarity between the left-out object and the other 9 objects used for training.

*Feature Projections in Embedding Spaces*

To generate putative ratings for objects along particular features using embedding spaces, we adapted and extended a previously used vector projection method[30]. We dub our method "contextual semantic projection" and it starts with manually defining three objects each that represent the extreme ends of a particular feature (e.g., for the "size" feature, objects representing the low end are 'bird', 'rabbit', and 'rat', and objects representing the high end are 'lion', 'giraffe', and 'elephant'). Subsequently, for each feature, nine vectors are defined in the embedding space as the vector differences between all possible pairs of an object representing the low extreme of a feature and an object representing the high extreme of a feature (e.g., the vector difference between word 'bird' and word 'lion', word 'rat' and word 'giraffe', etc.). The average of these nine vector differences represents a one-dimensional subspace of the original embedding space (line) and is used as approximation of its corresponding feature (e.g., the 'size' feature vector). To avoid overfitting, and given the high inter-rater reliability observed for test object feature ratings (r=0.68–0.92), the contextual projection endpoints for each feature were chosen by the experimenters as reasonable examples of out-of-sample objects representative of the low/high value on their corresponding feature (i.e., distinct from the 10 test objects used for each semantic context). All objects used as endpoints across each feature and semantic context are shown in Supplementary Table 5 (nature semantic context) and Supplementary Table 6 (transportation semantic context).

Once a feature subspace was defined, the rating of an object with respect to that feature was calculated by projecting the vector representing the object in the original embedding space onto the one-dimensional feature subspace, which resulted in a scalar value (overall range across all models, features, and contexts: [-0.6, 0.4]):

$$rating_{object} = \frac{feature^T object}{||feature||}$$

To illustrate the relationship with cosine distance in the original embedding space, we note that the difference between the feature ratings of two words is then equivalent to the normalized cosine distance between the vector difference of those two words in the original embedding space and the corresponding feature vector:

$$dist(object_1, object_2) = \frac{feature^T(object_1 - object_2)}{||feature||} =$$

$$= cosineDist(object_1 - object_2, feature) \cdot ||object_1 - object_2||$$

By contrast, prior work using projection techniques to extract feature ratings from embedding spaces[31] have used adjectives as endpoints, unconstrained by semantic context (e.g., 'size' was defined as a vector from 'small', 'tiny', 'minuscule' to 'large', 'huge', 'giant', regardless of context). To test the hypothesis that endpoints of a feature projection may benefit from semantic context constraints, we compared our semantic contextual projection technique to the adjective projection technique in terms of predicting empirical feature ratings. A complete list of the adjective projection endpoints used for each semantic context and each feature are listed in Supplementary Tables 5 & 6.

*Statistics*

For the non pre-trained models, we averaged over n=10 different learned embedding representations (10 different initial conditions) to obtain a mean similarity prediction for each type of model. All error bars reported were 95% confidence intervals using 1000 bootstrapped samples of the test-set items with replacement. For each model comparison in each condition, we used a non-parametric statistical significance estimation procedure to obtain p-values based on the aforementioned bootstrap sampling[66]: to compare two models, we sampled a correlation value from each one and computed the difference, repeating 1000 times, once for each bootstrapped sample to obtain a distribution of differences; we then estimated the p-value of the difference between the two models as 1 minus the proportion of values in this distribution that fell above zero. All correlation values reported are Pearson r correlation coefficients.

**Data and Code Availability**

Before publication, the data and the code for all experiments, models, and analyses that support the findings of this study are available from the corresponding author upon reasonable request. After publication, all data and code will immediately be made publicly available on the corresponding author's website and Github repository.

## Author Contributions

M.C.I. and J.D.C. designed the study. M.C.I. and T.G. collected and analyzed the data with input from C.T.E., N.M.B., and J.D.C. M.C.I., T.G., and J.D.C wrote the manuscript with input from C.T.E. and N.M.B.

## Competing Interests

The author(s) declare no competing financial interests.

**Supplementary Information for**

**Context Matters: Recovering Human Semantic Structure from Machine Learning Analysis of Large-Scale Text Corpora**

*Marius Cătălin Iordan\*, Tyler Giallanza, Cameron T. Ellis, Nicole M. Beckage, Jonathan D. Cohen*

**Supplementary Experiments 1–4**

*Methods*

This section assumes familiarity with the main text Methods section. To test whether similarity judgments are influenced by the semantic context in which they are made[38,39,40,41,42] for our test objects, we ran four behavioral experiments, where we manipulated the semantic context of similarity judgments by titrating the amount of judgments from the animal and vehicle semantic contexts that the participants were asked to perform.

To collect empirical similarity judgments, we recruited 665 participants (228 female, 397 right-handed, mean age 37.2 years) through the Amazon Mechanical Turk online platform in exchange for $1.25 payment (expected rate $7.50/hour). Participants were asked to report the similarity between 45 pairs of objects on a discrete scale of 1 to 5 (1 = not similar; 5 = very similar). In each trial, the participant was shown two randomly selected images from the same semantic category side-by-side and was given unlimited time to report a similarity judgment. In Supplementary Experiment 1 (50%-50% titration), participants performed 22 judgments between randomly selected pairs of objects from one semantic context (e.g., nature) and 23 judgments from the other semantic context (e.g., transportation). In Supplementary Experiments 2–3 (70%-30%), participants performed 31 judgments between randomly selected pairs of objects from one semantic context (e.g., nature) and 14 judgments from the other semantic context (e.g., transportation), and vice-versa. In Supplementary Experiment 4 (90%-10%), participants performed 40 judgments between randomly selected pairs of objects from one semantic context (e.g., nature) and 5 judgments from the other semantic context (e.g., transportation). For the latter Experiment, we only included one direction (90% nature – 10% transportation) due to the prohibitive cost associated with running such an experiment with reliable results for the 10% context. Since trials from each semantic context were presented randomly, we ensured that at least 20 unique trials per comparison were collected in all Experiments.

To ensure high-quality judgments, we limited participation only to Mechanical Turk workers who had previously completed at least 1000 HITs with an acceptance rate of 95% or above. We excluded 63 participants who had no variance across answers (e.g. choosing a similarity value of 1 for every object pair). Prior work has shown that for this type of task inter-participant reliability should be high[23], therefore to exclude participants whose response may have been random, we correlated the responses of each participant with the average of the responses for every other participant and calculated the Pearson correlation coefficient. We then iteratively removed the participant with the lowest Pearson coefficient, stopping this procedure when all remaining participants had a Pearson coefficient greater than or equal to 0.5 to the rest of the group. This same method was applied to behavioral results mentioned in the main text. This method excluded an additional 208 participants, leading to a final tally of n=394 participants across all 4 Supplementary Experiments.

*Results*

We compared the similarity patterns obtained in Supplementary Experiments 1–4 to those in Experiment 1 in the main text using Pearson correlation. As predicted, we observed a gradual decrease in the amount of agreement between the Supplementary similarity judgments and the Experiment 1 judgments as the mix of judgments from the two contexts became increasingly asymmetric – Supplementary Experiments 1–4 (Supplementary Fig. 1; p<0.001 for both contexts, Friedman test). Consistent with prior work[38,39,40,41,42], our results indicate that semantic contextual constraints influence empirical semantic similarity judgments.

**Supplementary Experiment 5**

*Methods*

This section assumes familiarity with the main text Methods section. To show that the difference in performance between our contextually-constrained embedding models on their respective test sets cannot be explained by the word frequency alone and/or the absolute number of test words appearing in the training corpora, we constructed new training corpora for our two semantic contexts that matched both the number of occurrences and the word frequency of the test items (animals and vehicles) across both corpora. For the nature context, we sought to build a nature semantic corpus that matched both the frequency and quantity of test animals to that of the transportation corpus. Thus, we computed the total number of occurrences of individual test animals in the transportation corpus and sub-sampled from the original nature corpus enough articles that contained the same number of each test animal as the transportation corpus, as well as additional articles that contained no animals up to the maximum size of corpus we could build this way from the remaining articles (51M words; Supplementary Table 2). Similarly, for the transportation context, we sought to build a transportation semantic corpus that matched both the frequency and quantity of test vehicles to that of the nature corpus. Thus, we computed the total number of occurrences of individual test vehicles in the nature corpus and sub-sampled from the original transportation corpus enough articles that contained the same number of each test vehicle as the nature corpus, as well as additional articles that contained no vehicles up to the maximum size of corpus we could build this way from the remaining articles (38M words; Supplementary Table 2).

Using these new corpora, we would expect a decrease in performance compared to the original (context-specific) corpora as we are removing exemplars that are extremely relevant to the similarity task. However, we also believe that even though the target words are seen much less often, the corpus trained in the proper context would still outperform the general corpus or the corpus trained in an 'incorrect' context. To test this hypothesis, we then trained new word embeddings using a frequency-matched corpus using the same methodology as in Experiment 1 using the new corpora. After retraining, we tested how well the new models could predict empirical similarity judgments using the same procedure as in Experiment 1 (see main Methods and Results).

*Results*

Using the frequency-matched models, we observed the same double dissociation between the ability of (matched) contextually-constrained embeddings to predict human similarity judgments better for their preferred context, compared to their non-preferred context (Supplementary Fig. 7; nature context: contextually-constrained nature 51M frequency-matched $r=0.420\pm0.002$(CI), contextually-constrained transportation 51M frequency-matched $r=0.180\pm0.003$(CI); $p<0.001$;

transportation context: contextually-constrained transportation 38M frequency-matched r=0.520±0.003(CI), contextually-constrained nature 38M frequency-matched r=0.380±0.004(CI); p<0.001). Together, our results suggest that any discrepancies in word frequency and absolute number of test words in the training corpus cannot fully explain the performance of our contextually-constrained models above context-free models.

***Supplementary Figure 1. Semantic Context Influences Empirical Similarity Judgments.*** *In Supplementary Experiments 1–4, we titrated the amount of similarity judgments participants had to perform from our two separate semantic contexts: e.g., 70% nature – 30% transportation (see Supplementary Methods). We found that the fewer judgments a participant had to make from a given semantic context, the less correlated those judgments were to the original empirical results from Experiment 1 where all judgments were done within the same semantic context (100% bar below).*

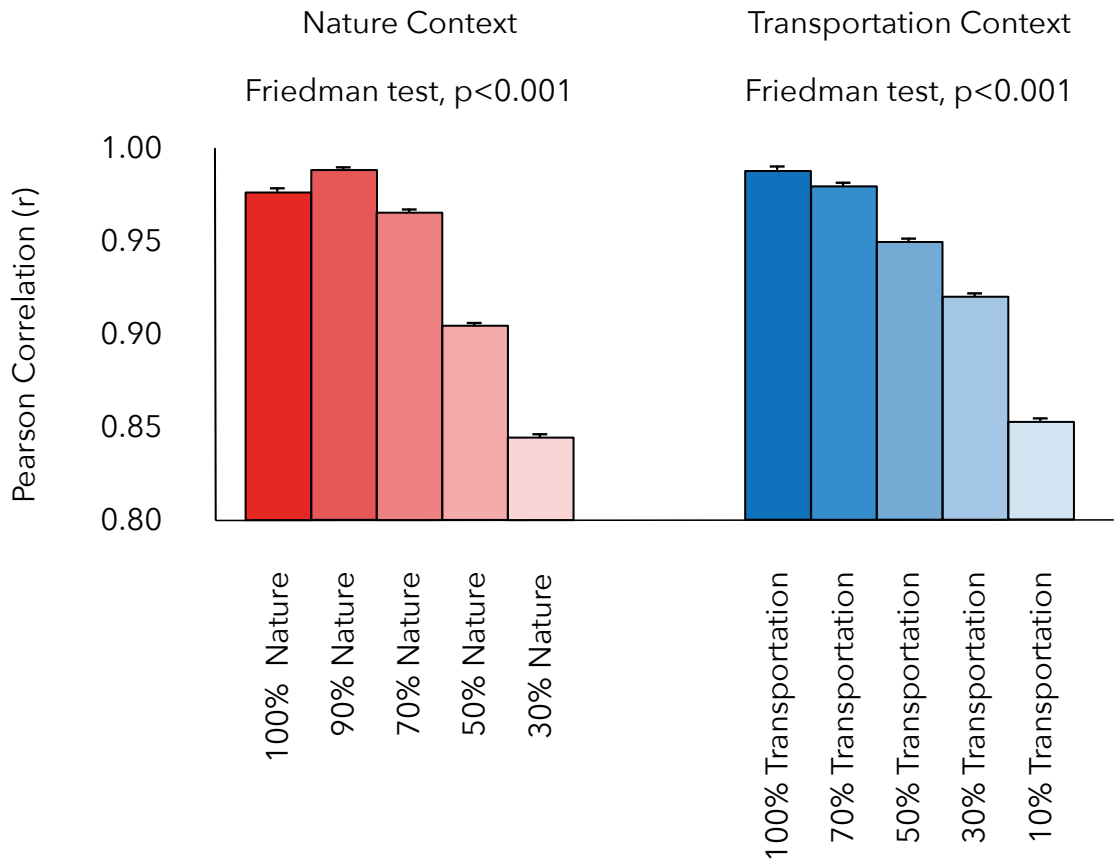

***Supplementary Figure 2. Nature Context – Correlation Between Human Similarity Judgments and Context-Free Embedding Spaces for Varying Word2Vec Model Training Parameters: Syntactic Word Window Size 8–12, Dimensionality 100-200.*** *Pearson correlation between object similarity predictions (via cosine distances between pairs of animals in each embedding space) and human similarity judgments.*

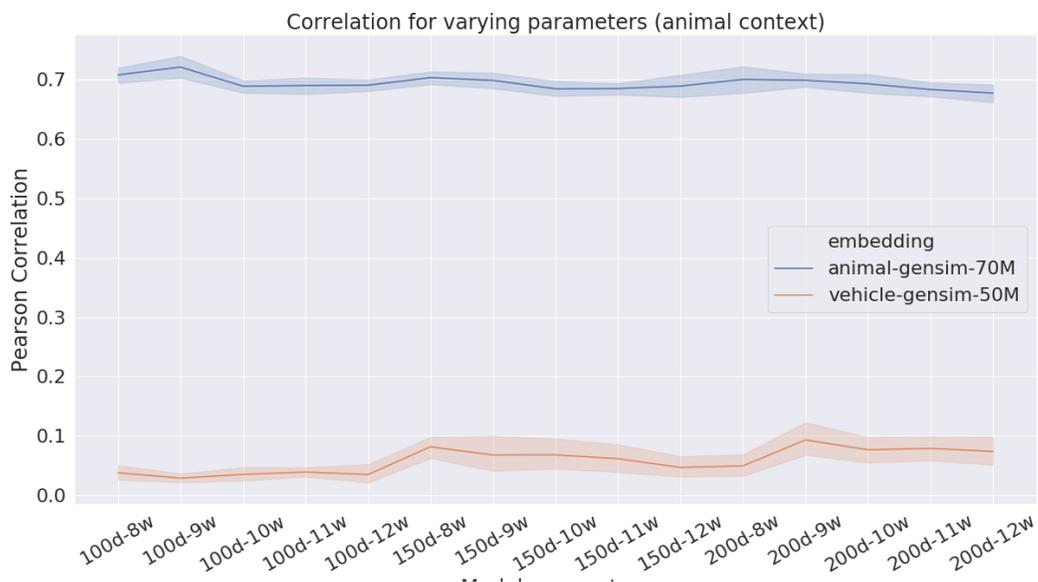

*Supplementary Figure 3. Transportation Context – Correlation Between Human Similarity Judgments and Context-Free Embedding Spaces for Varying Word2Vec Model Training Parameters: Syntactic Word Window Size 8–12, Dimensionality 100-200.* Pearson correlation between object similarity predictions (via cosine distances between pairs of vehicles in each embedding space) and human similarity judgments.

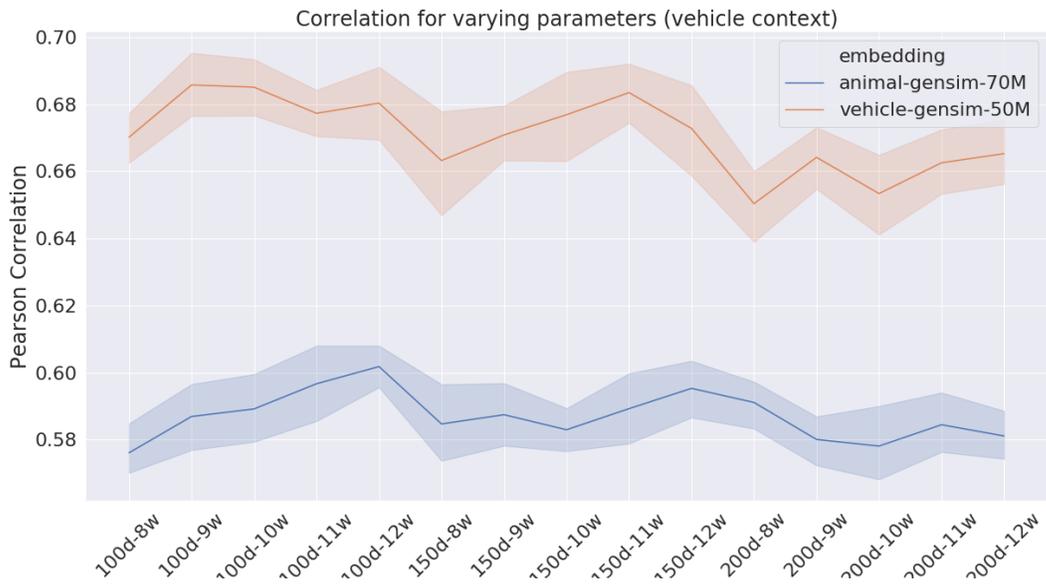

***Supplementary Figure 4. Number of Initializations for Each Model Does Not Change Prediction Results.*** *To ensure that the difference in performance between embedding models reported in the manuscript is not due to training only 10 models for each condition, we selected a subset of models in Fig. 1 (Contextually-Constrained Nature, Contextually-Constrained Transportation, Context-Free Wikipedia) and ran 50 models using different random initializations for each type of model. We found that running more models per condition did not change performance for contextually-constrained embeddings, and slightly reduced performance for context-free embeddings in the Nature semantic context.*

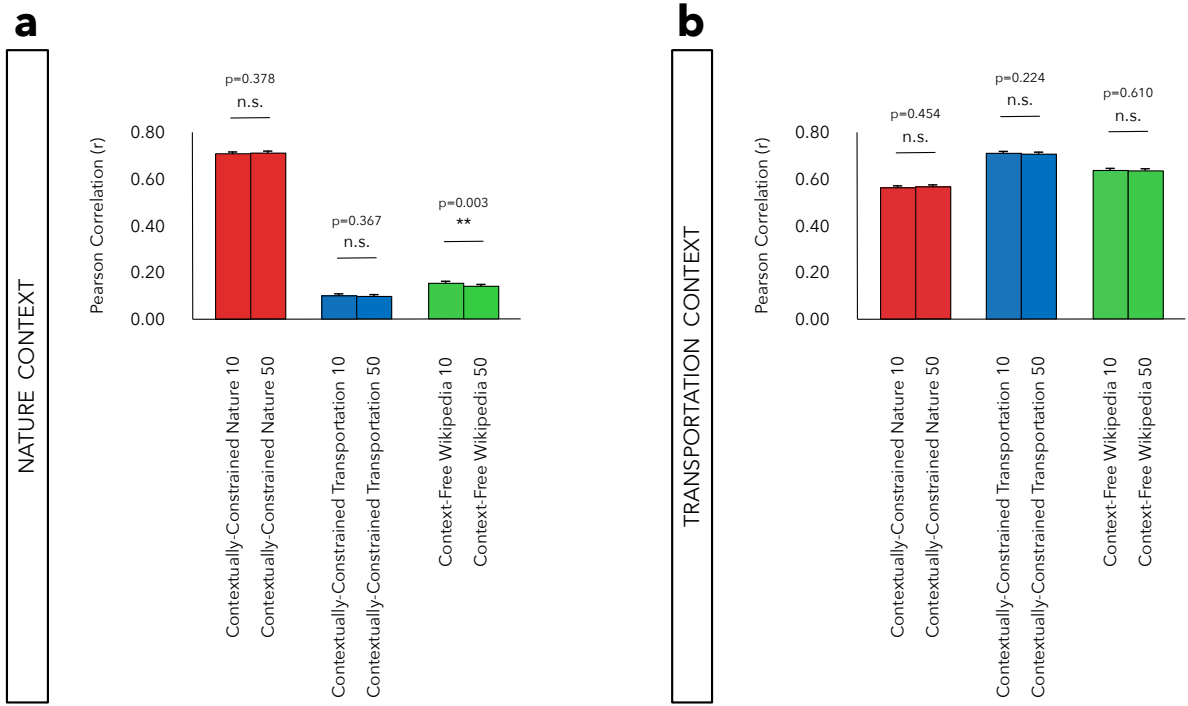

***Supplementary Figure 5. Using Spearman Instead of Pearson Correlation Does Not Alter The Performance Gap Between Contextually-Constrained and Context-Free Embedding Models.*** To account for potential non-linear effects in matching predictions of the embedding models with empirical similarity judgments, we used Spearman correlation to re-evaluate the relationships between the models in Fig. 1 and human data. We found no qualitative difference in the relative pattern of effects across our models.

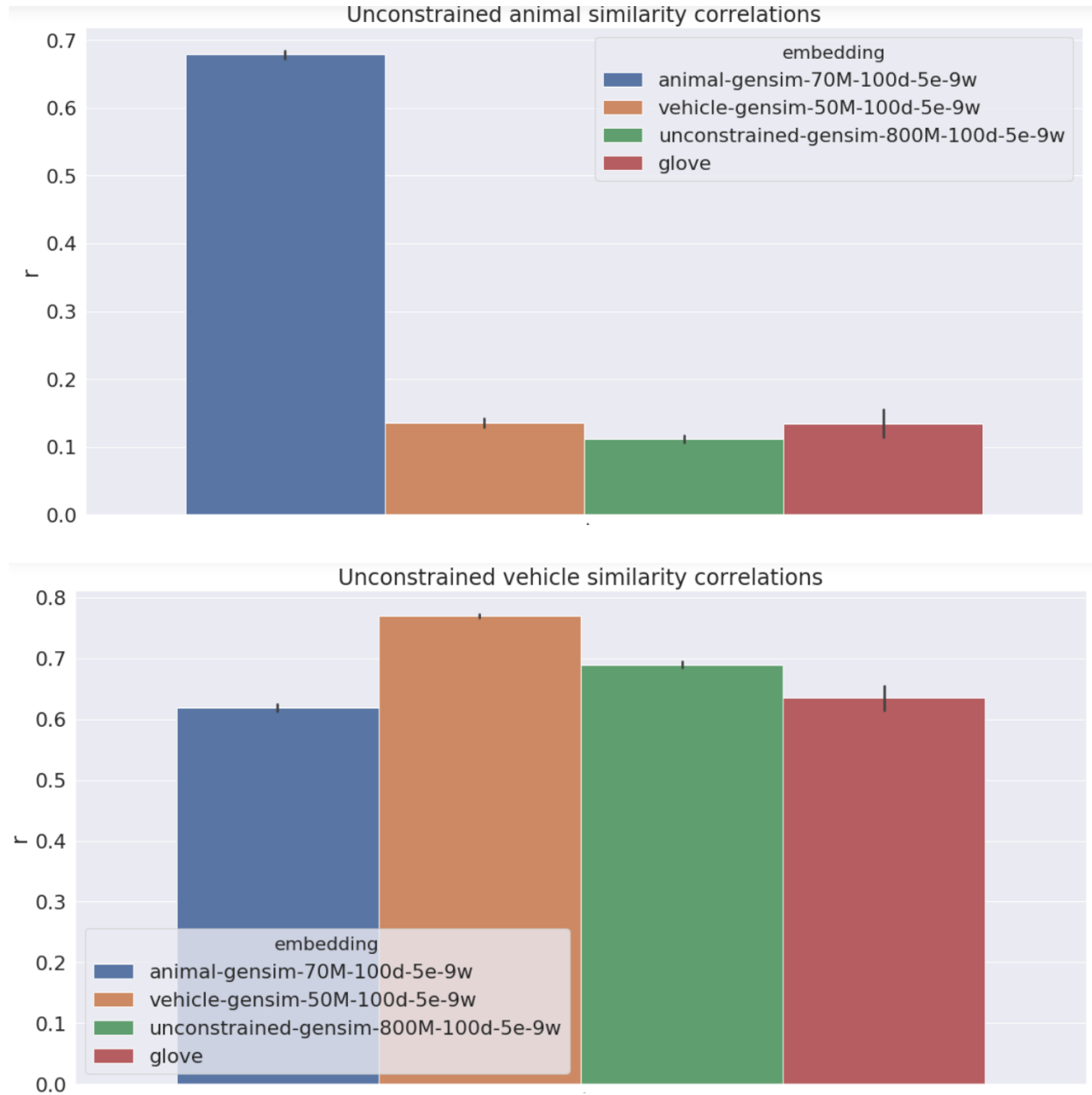

***Supplementary Figure 6. Similarity Matrices for Main Models in Fig. 1.*** In Fig. 1, we reported the Pearson correlation between similarity predictions made by artificial neural networks and human empirical data. We include below the similarity matrices corresponding to the human data and the three major embedding models we trained: contextually-constrained Nature, contextually-constrained Transportation, and context-free Wikipedia. We did not observe any obvious trends in the errors or agreements made by networks when compared to human judgments. Additionally, our bootstrapping procedure used to report statistics in the main text precludes significant influences by item-level effects.

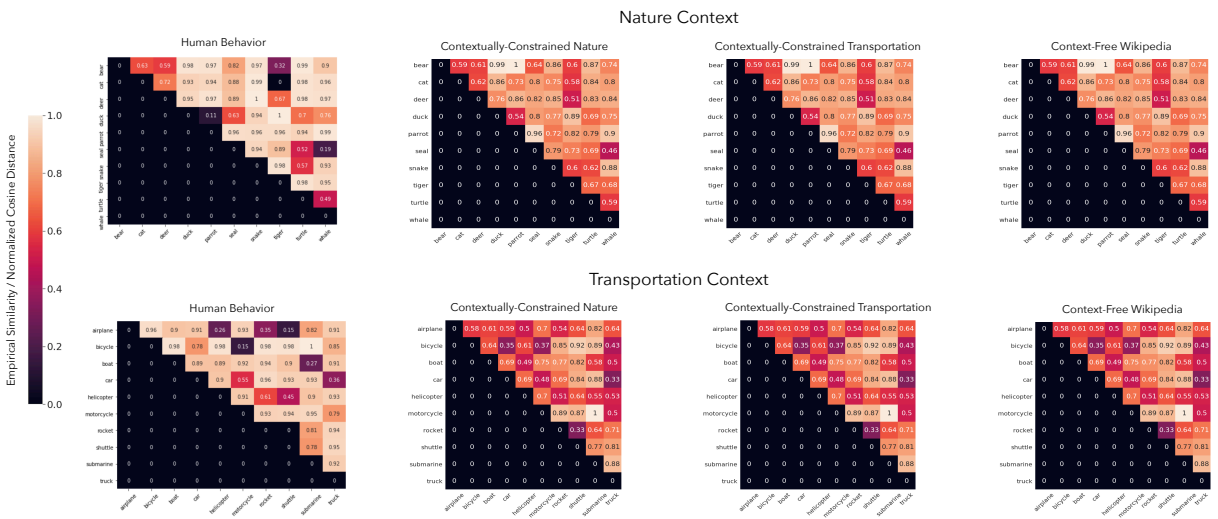

***Supplementary Figure 7. Word Frequency Differences for Animals and Vehicles Between Training Corpora Do Not Influence Relative Performance of Models.*** We constructed new training corpora for our two semantic contexts that matched both the number of occurrences and the word frequency of the test items (animals and vehicles) across both corpora (i.e., two training corpora of 50M words each containing the same number of animals and vehicles each). After re-training our models, we observed the same double dissociation between the ability of (matched) contextually-constrained embeddings to predict human similarity judgments better for their preferred context, compared to their non-preferred context.

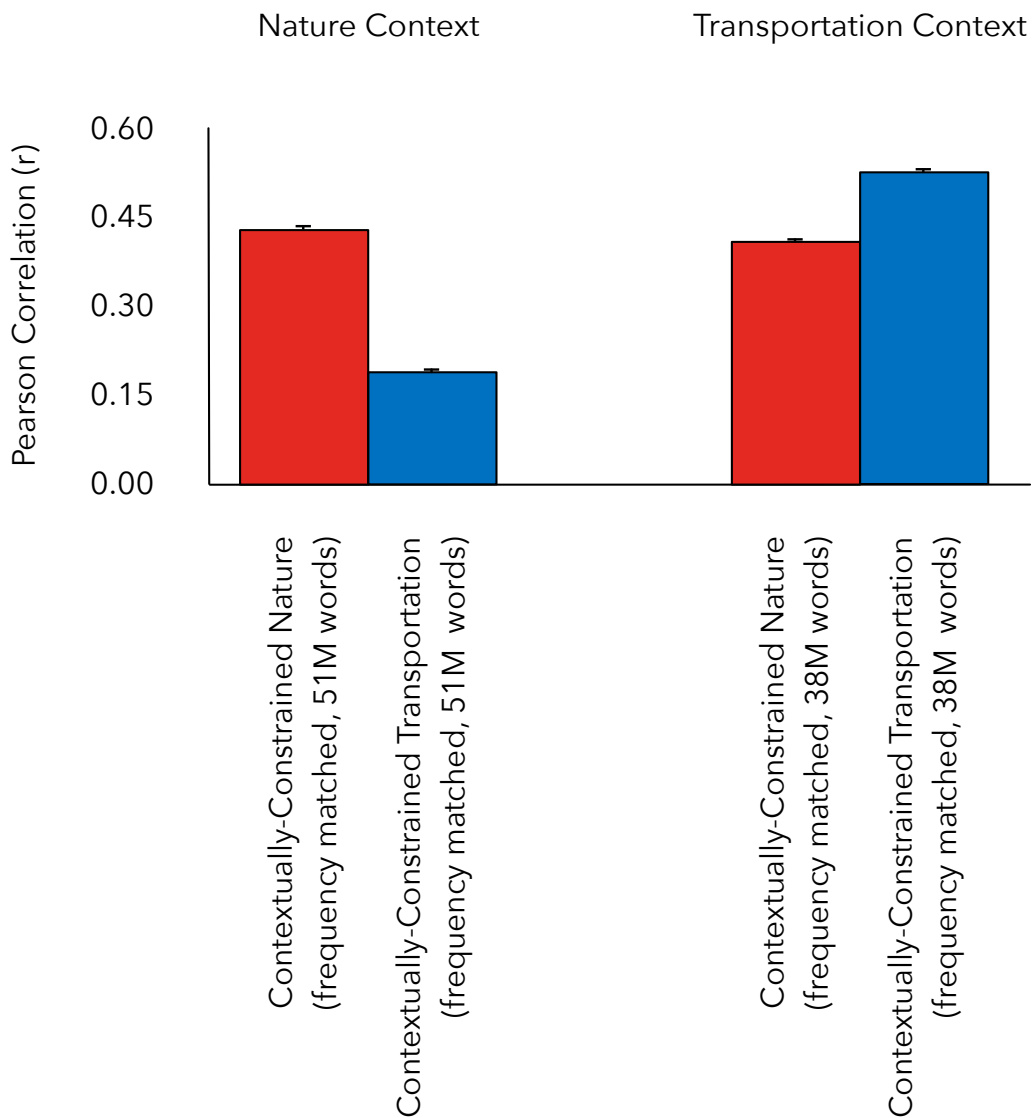

***Supplementary Figure 8. Confidence Intervals for Feature Ratings Predictions.*** To show that our models' ability to predict empirical feature ratings in Fig. 3 is not due to initialization conditions or item-level effects, we report below the 95% confidence intervals for 10 independent initializations of each model and 1000 bootstrapped samples of the test-set items per model.

### Nature Context

| Dimension / Model | Aquaticness | Cuteness | Dangerousness | Domesticity | Edibility | Furriness | Humanness | Intelligence | Interestingness | Predacity | Size | Speed |
|---|---|---|---|---|---|---|---|---|---|---|---|---|
| **Contextual Projection** | | | | | | | | | | | | |
| Nature | **.90 ± .00** | .55 ± .01 | .58 ± .01 | **.66 ± .01** | **.75 ± .02** | **.69 ± .01** | **.72 ± .01** | **.62 ± .02** | .48 ± .02 | **.69 ± .01** | **.93 ± .01** | .60 ± .02 |
| Transportation | .77 ± .01 | .46 ± .02 | .53 ± .02 | .31 ± .01 | .45 ± .01 | .21 ± .01 | .30 ± .01 | .34 ± .01 | **.50 ± .02** | .49 ± .01 | .46 ± .01 | **.61 ± .01** |
| Wikipedia (full) | .74 ± .01 | **.77 ± .01** | .52 ± .02 | .40 ± .02 | .72 ± .01 | .63 ± .01 | .49 ± .01 | .29 ± .01 | .32 ± .02 | .60 ± .01 | .60 ± .01 | **.63 ± .01** |
| **Adjective Projection** | | | | | | | | | | | | |
| Nature | .81 ± .01 | .51 ± .01 | .37 ± .01 | **.66 ± .02** | .56 ± .01 | .61 ± .01 | .48 ± .01 | .36 ± .01 | .41 ± .01 | .56 ± .01 | .56 ± .01 | .51 ± .01 |
| Transportation | .66 ± .01 | .26 ± .02 | .36 ± .01 | .50 ± .02 | .48 ± .01 | .29 ± .02 | .28 ± .01 | **.60 ± .01** | **.49 ± .02** | .41 ± .01 | .41 ± .01 | .29 ± .01 |
| Wikipedia (full) | .46 ± .02 | .48 ± .02 | **.68 ± .01** | .33 ± .03 | .22 ± .02 | .26 ± .01 | .25 ± .01 | .37 ± .02 | .46 ± .01 | .46 ± .02 | .47 ± .02 | .29 ± .01 |

### Transportation Context

| Dimension / Model | Comfort | Cost | Dangerousness | Elevation | Interestingness | Openness | Personalness | Size | Skill | Speed | Usefulness | Wheeledness |
|---|---|---|---|---|---|---|---|---|---|---|---|---|
| **Contextual Projection** | | | | | | | | | | | | |
| Nature | .37 ± .01 | .89 ± .02 | .80 ± .02 | .61 ± .02 | **.81 ± .02** | **.78 ± .02** | **.88 ± .02** | **.82 ± .02** | .85 ± .01 | .59 ± .02 | .36 ± .02 | .50 ± .02 |
| Transportation | **.48 ± .01** | **.94 ± .01** | **.84 ± .01** | **.75 ± .01** | **.80 ± .02** | **.81 ± .03** | .71 ± .01 | **.85 ± .03** | **.90 ± .02** | **.71 ± .01** | **.39 ± .01** | **.77 ± .02** |
| Wikipedia (full) | .41 ± .01 | .92 ± .01 | .72 ± .01 | .59 ± .01 | .77 ± .01 | .70 ± .02 | **.89 ± .01** | .81 ± .02 | **.88 ± .02** | .54 ± .01 | .36 ± .02 | .73 ± .01 |
| **Adjective Projection** | | | | | | | | | | | | |
| Nature | **.48 ± .02** | .45 ± .02 | .34 ± .02 | .40 ± .02 | .51 ± .01 | .28 ± .01 | .51 ± .01 | .49 ± .01 | .64 ± .01 | .31 ± .01 | .28 ± .02 | .73 ± .02 |
| Transportation | .35 ± .02 | .52 ± .01 | .35 ± .00 | .32 ± .02 | .27 ± .02 | .31 ± .01 | .48 ± .01 | .36 ± .01 | .45 ± .02 | .65 ± .02 | .33 ± .01 | **.75 ± .02** |
| Wikipedia (full) | .27 ± .01 | .48 ± .01 | .61 ± .02 | .35 ± .01 | .41 ± .01 | .35 ± .02 | .57 ± .01 | .33 ± .02 | .80 ± .02 | .55 ± .02 | .32 ± .03 | .71 ± .02 |

***Supplementary Figure 9. Confidence Intervals for Feature Rating Predictions for the Combined-Context Models.*** To show that our models' ability to predict empirical feature ratings in Fig. 4 is not due to initialization conditions or item-level effects, we report below the 95% confidence intervals for 10 independent initializations of each model and 1000 bootstrapped samples of the test-set items per model.

Nature Context

| Dimension / Model | Aquaticness | Cuteness | Dangerousness | Domesticity | Edibility | Furriness | Humanness | Intelligence | Interest | Predacy | Size | Speed |
|---|---|---|---|---|---|---|---|---|---|---|---|---|
| Contextual Projection | | | | | | | | | | | | |
| **Nature** | **.90 ± .00** | .55 ± .01 | **.58 ± .01** | **.66 ± .01** | **.75 ± .02** | **.69 ± .01** | .72 ± .01 | **.62 ± .02** | **.48 ± .02** | **.69 ± .01** | **.93 ± .01** | .60 ± .02 |
| **Combined** | .80 ± .01 | .52 ± .00 | .52 ± .00 | .29 ± .01 | .70 ± .01 | .62 ± .01 | **.77 ± .01** | **.60 ± .02** | .45 ± .01 | .58 ± .01 | .65 ± .01 | **.75 ± .01** |
| **Transportation** | .77 ± .01 | .46 ± .02 | .53 ± .02 | .31 ± .01 | .45 ± .01 | .21 ± .01 | .30 ± .01 | .34 ± .01 | **.50 ± .02** | .49 ± .01 | .46 ± .01 | .61 ± .01 |
| **Wikipedia (full)** | .74 ± .01 | **.77 ± .01** | .52 ± .02 | .40 ± .02 | .72 ± .01 | .63 ± .01 | .49 ± .01 | .29 ± .01 | .32 ± .02 | .60 ± .01 | .60 ± .01 | .63 ± .01 |

Transportation Context

| Dimension / Model | Comfort | Cost | Dangerousness | Elevation | Interest | Openness | Personalness | Size | Skill | Speed | Usefulness | Wheeledness |
|---|---|---|---|---|---|---|---|---|---|---|---|---|
| Contextual Projection | | | | | | | | | | | | |
| **Nature** | .37 ± .01 | .89 ± .02 | .80 ± .02 | .61 ± .02 | **.81 ± .02** | **.78 ± .02** | **.88 ± .02** | **.82 ± .02** | .85 ± .01 | .59 ± .02 | .36 ± .02 | .50 ± .02 |
| **Combined** | **.59 ± .01** | .91 ± .01 | .81 ± .01 | .61 ± .01 | **.81 ± .02** | .70 ± .01 | .86 ± .01 | .81 ± .01 | **.88 ± .02** | **.70 ± .01** | .35 ± .01 | .74 ± .01 |
| **Transportation** | .48 ± .01 | **.94 ± .01** | **.84 ± .01** | **.75 ± .01** | .80 ± .02 | **.81 ± .03** | .71 ± .01 | **.85 ± .03** | **.90 ± .02** | **.71 ± .01** | **.39 ± .01** | **.77 ± .02** |
| **Wikipedia (full)** | .41 ± .01 | .92 ± .01 | .72 ± .01 | .59 ± .01 | .77 ± .01 | .70 ± .02 | **.89 ± .01** | .81 ± .02 | **.88 ± .02** | .54 ± .01 | .36 ± .02 | .73 ± .01 |

***Supplementary Figure 10. Contextual Projection and Regression Performs Better Than Using Regression Using the Original Embedding Space.*** *We used linear regression to learn optimal weights on the original 100 dimensions of each embedding space (using a subset of the behavioral data) that best predicted human similarity judgments and used those weights to perform out-of-sample prediction on the remaining behavioral data. We found that regression alone cannot learn an alignment to predict human judgments from the high-dimensional embedding spaces we tested.*

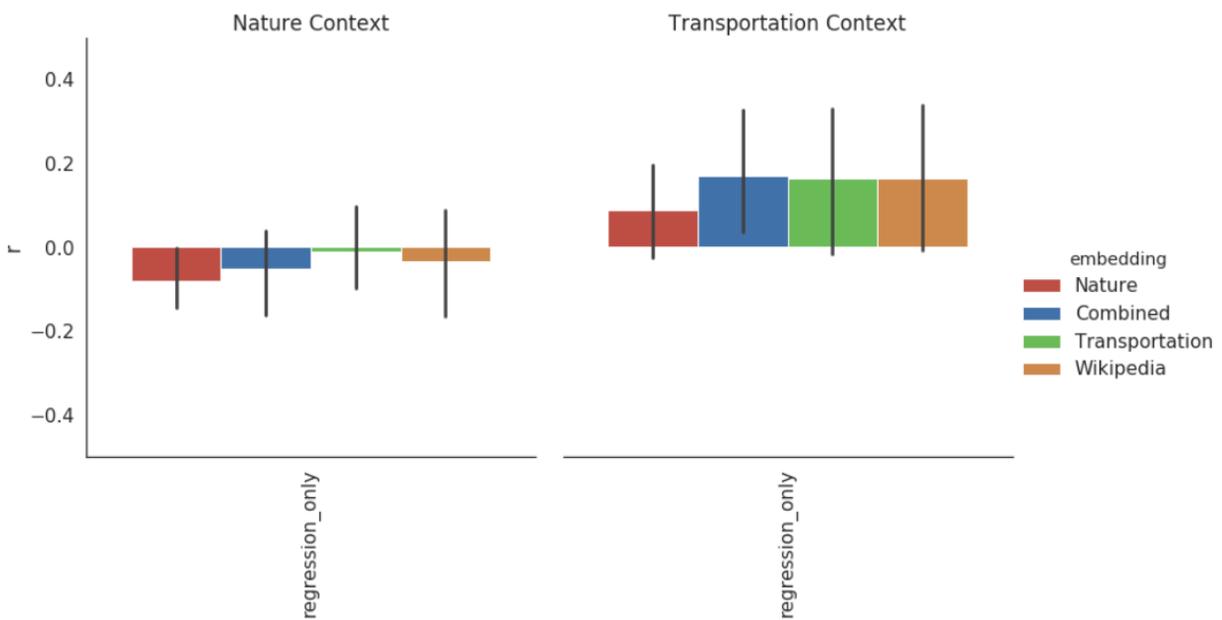

***Supplementary Figure 11. Contextual Projection and Regression Performs Better Than Using Cosine Distance in the 12-Dimensional Contextual Projection Space.*** *We used cosine distance to generate a similarity prediction for all pairs of test objects from the 12-dimensional subspaces obtained via contextual projection for each embedding space. We then used Pearson correlation (variance explained reported below) to measure how similar these predictions were to human judgments. We found that using cosine distance on the reduced-dimensionality contextual-projection subspaces alone cannot predict human judgments from the high-dimensional embedding spaces we tested as well as linear regression on the same dimensionality-reduced subspaces.*

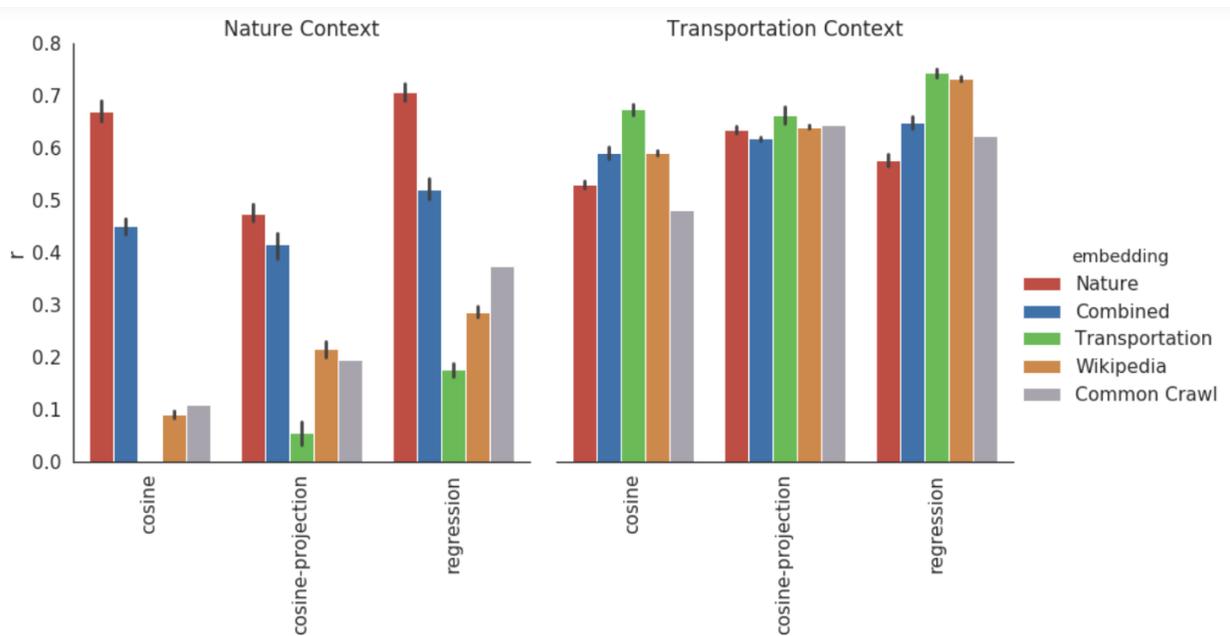

*Supplementary Table 1. Number of Meanings for Test Objects in Both Semantic Contexts (OED: Oxford English Dictionary).*

| Context | Object | OED Meanings | WordNet Meanings |
|---|---|---|---|
| Nature | Bear | 5 | 15 |
| | Cat | 8 | 10 |
| | Deer | 1 | 1 |
| | Duck | 5 | 8 |
| | Parrot | 3 | 3 |
| | Seal | 8 | 15 |
| | Snake | 3 | 9 |
| | Tiger | 2 | 2 |
| | Turtle | 4 | 4 |
| | Whale | 3 | 3 |
| | 95 % Confidence Interval | 3.0 – 5.4 | 4.4 – 9.6 |
| Transportation | Airplane | 1 | 1 |
| | Boat | 3 | 3 |
| | Bicycle | 2 | 2 |
| | Car | 5 | 5 |
| | Helicopter | 2 | 1 |
| | Motorcycle | 2 | 2 |
| | Rocket | 7 | 7 |
| | Shuttle | 6 | 4 |
| | Submarine | 2 | 7 |
| | Truck | 7 | 3 |
| | 95 % Confidence Interval | 3.6 – 4.8 | 2.4 – 4.6 |
| Statistics | **Comparison** | **T(9)** | **p-value** |
| | OED Nature vs. Transportation | 0.480 | 0.637 |
| | WordNet Nature vs. Transportation | 1.955 | 0.066 |

*Supplementary Table 2. Number of Occurrences of Test Objects in Frequency-Matched Text Corpora (Supplementary Experiment 5 & Supplementary Fig. 7).*

| Object Context | Object | Nature-Matched (51M) | Transportation-Matched (38M) |
|---|---|---|---|
| Nature | Bear | 1,227 | 1,697 |
| | Cat | 1,646 | 1,120 |
| | Deer | 349 | 657 |
| | Duck | 315 | 297 |
| | Parrot | 21 | 91 |
| | Seal | 1,252 | 1,003 |
| | Snake | 231 | 767 |
| | Tiger | 1,009 | 530 |
| | Turtle | 114 | 653 |
| | Whale | 204 | 1,123 |
| Transportation | Airplane | 478 | 544 |
| | Boat | 1,274 | 1,892 |
| | Bicycle | 513 | 645 |
| | Car | 3,006 | 4,419 |
| | Helicopter | 808 | 1,085 |
| | Motorcycle | 129 | 156 |
| | Rocket | 2,502 | 3,640 |
| | Shuttle | 682 | 1,683 |
| | Submarine | 802 | 975 |
| | Truck | 596 | 777 |

***Supplementary Table 3. List of Features Selected for Nature Semantic Context.***

| Feature | Definition | Participants |
|---|---|---|
| Size | An animal's overall dimensions; how big or small an animal is. | 23 |
| Domesticity | How tame an animal is; the likelihood the animal would be kept by humans. | 25 |
| Predacity | An animal's tendency to hunt other animals; an animal could be prey, predator, or in-between. | 23 |
| Speed | How fast or slow an animal moves in its natural habitat; the animal's potential to move fast or slow. | 22 |
| Furriness | Whether an animal's coat or skin is covered in fur, hair, or feathers; related to an animal's fluffyness. | 25 |
| Aquatic-ness | An animal's tendency to live eat, hunt, or grow in water (lake, sea, river, ocean), compared to on land; the amount of time the animal spends in contact with water. | 24 |
| Dangerousness | How dangerous would an encounter with this animal be on average; the likelihood of the animal hurting a human on contact. | 23 |
| Edibility | How fit the animal is for human consumption; is the animal usually eaten by humans. | 25 |
| Intelligence | How smart the animal is; the degree of intelligence that the animal possesses. | 23 |
| Humanness | How many characteristics the animal has in common with humans; how easy it is to attribute human characteristics to the animal. | 21 |
| Cuteness | How attractive or adorable the animal is. | 33 |
| Interest | How interesting you find the animal; how much curiosity or attention the animal would draw when seen. | 23 |

## Supplementary Table 4. List of Features Selected for Transportation Semantic Context.

| Feature | Definition | Participants |
|---|---|---|
| Size | A vehicle's overall dimensions; how big or small a vehicle is. | 32 |
| Cost | The average cost of purchasing a vehicle. | 24 |
| Open-ness | How open the vehicle is to the outside world; the level of protection the vehicle gives against the elements. | 23 |
| Speed | How fast or slow a vehicle moves during normal operation; the vehicle's potential to move fast or slow. | 29 |
| Wheeled-ness | How often the vehicle utilizes wheels during normal operation; the degree to which you associate the vehicle with using wheels. | 24 |
| Dangerousness | How dangerous would using this vehicle be on average; the likelihood of the vehicle's driver and passengers getting hurt while operating the vehicle. | 26 |
| Elevation | The elevation (underwater, ground-level, in the sky, in space) at which the vehicle normally operates; how high above or below the ground the vehicle typically is. | 23 |
| Comfort | How comfortable the vehicle is for its passengers and operator. | 23 |
| Skill | How much skill is required to operate the vehicle; the level of training required before being capable of operating the vehicle. | 22 |
| Personal-ness | How much personal attachment the operator of the vehicle would have towards the vehicle; how often the vehicle is operated by its owner. | 24 |
| Useful-ness | How useful you find the vehicle; how capable the vehicle is of performing a valuable function. | 18 |
| Interest | How interesting you find the vehicle; how much curiosity or attention the vehicle would draw when seen. | 23 |

*Supplementary Table 5. Feature Endpoints for Adjective and Contextual Projection Methods – Nature Semantic Context.*

| Feature | Low-End Adjectives | High-End Adjectives | Low-End Objects | High-End Objects |
|---|---|---|---|---|
| Size | {small, little, tiny} | {big, large, huge} | {bird, rabbit, rat} | {lion, giraffe, elephant} |
| Domesticity | {wild, untamed, undomesticated} | {domestic, pet, tamed} | {fish, pig, ferret} | {shark, lion, panther} |
| Predacity | {quarry, prey, herbivore} | {ferocious, predatory, carnivorous} | {rabbit, goose, swan} | {shark, lion, wolf} |
| Speed | {slow, sluggish, gradual} | {fast, quick, speedy} | {sloth, snail, tortoise} | {leopard, cheetah, hummingbird} |
| Furriness | {smooth, sleek, rough} | {fast, quick, speedy} | {penguin, dolphin, frog} | {puma, dog, lion} |
| Aquatic-ness | {terrestrial, land, walking} | {water, aquatic, swimming} | {lion, panther, dog} | {dolphin, fish, frog} |
| Dangerousness | {dangerous, unsafe, threatening} | {safe, harmless, innocuous} | {swan, rabbit, goose} | {shark, lion, scorpion} |
| Edibility | {uneatable, inedible, indigestible} | {delicious, edible, food} | {dog, lion, elephant} | {cow, veal, chicken} |
| Intelligence | {dumb, stupid, idiotic} | {smart, intelligent, wise} | {lizard, goose, snail} | {crow, elephant, dolphin} |
| Humanness | {inhuman, animal, wild} | {human, anthropomorphic, humanist} | {goose, swan, frog} | {dog, ape, monkey} |
| Cuteness | {ugly, repulsive, hideous} | {cute, adorable, attractive} | {shark, raccoon, lizard} | {rabbit, dog, giraffe} |
| Interest | {uninteresting, lame, boring} | {interesting, cool, exciting} | {goose, pigeon, snail} | {gorilla, dolphin, lion} |

*Supplementary Table 6. Feature Endpoints for Adjective and Contextual Projection Methods – Transportation Semantic Context.*

| Feature | Low-End Adjectives | High-End Adjectives | Low-End Objects | High-End Objects |
|---|---|---|---|---|
| Size | {small, little, tiny} | {big, large, huge} | {segway, scooter, skateboard} | {spaceship, carrier, airliner} |
| Cost | {cheap, affordable, inexpensive} | {costly, expensive, pricey} | {bike, scooter, skateboard} | {fighter, spaceship, carrier} |
| Openness | {covered, windowless, enclosed} | {uncovered, open, exposed} | {spaceship, jet, tank} | {convertible, skateboard, scooter} |
| Speed | {slow, sluggish, gradual} | {fast, quick, speedy} | {van, barge, bus} | {speedboat, racecar, spaceship} |
| Wheeledness | {hull, smooth, propelled} | {wheeled, wheels, tire} | {kayak, speedboat, spaceship} | {convertible, racecar, bus} |
| Dangerousness | {dangerous, unsafe, threatening} | {safe, harmless, innocuous} | {canoe, cart, buggy} | {spaceship, destroyer, jet} |
| Elevation | {underwater, low, deep} | {high, elevated, skyward} | {automobile, convertible, yacht} | {spaceship, satellite, jet} |
| Comfort | {uncomfortable, cramped, hunched} | {comfortable, cozy, relaxing} | {scooter, cart, tractor} | {sedan, yacht, train} |
| Skill | {simple, novice, unskilled} | {complex, skilled, experienced} | {tricycle, bus, train} | {aircraft, spaceship, jet} |
| Personalness | {impersonal, public, shared} | {personal, intimate, private} | {airliner, spaceship, carrier} | {skateboard, sportscar, yacht} |
| Usefulness | {uncommon, useless, novelty} | {useful, usable, common} | {skateboard, yacht, jetpack} | {airliner, bus, train} |
| Interest | {uninteresting, lame, boring} | {interesting, cool, exciting} | {scooter, minivan, train} | {jet, spaceship, yacht} |